\documentclass[sigconf]{acmart}

\AtBeginDocument{%
  \providecommand\BibTeX{{%
    \normalfont B\kern-0.5em{\scshape i\kern-0.25em b}\kern-0.8em\TeX}}}

\setcopyright{acmlicensed}
\copyrightyear{2018}
\acmYear{2018}
\acmDOI{XXXXXXX.XXXXXXX}

\acmConference[Conference acronym 'XX]{Make sure to enter the correct
  conference title from your rights confirmation emai}{June 03--05,
  2018}{Woodstock, NY}
\acmISBN{978-1-4503-XXXX-X/18/06}

\usepackage{multirow}
\usepackage{multicol}
\usepackage{float}
\usepackage{microtype}
\usepackage{graphicx}
\usepackage{subfigure}
\usepackage{booktabs} 

\usepackage{wrapfig}

\usepackage{colortbl} 
\usepackage{xcolor} 

\usepackage{amsthm}

\theoremstyle{plain}
\newtheorem{theorem}{Theorem}

\usepackage{graphicx} 

\begin{document}

\title{Subgraph-level Universal Prompt Tuning}

\author{Junhyun Lee}
\email{ljhyun33@korea.ac.kr}
\affiliation{%
  \institution{Korea University}
  \city{Seoul}
  \country{South Korea}
  \postcode{02841}
}

\author{Wooseong Yang}
\email{yus0363@gmail.com}
\affiliation{%
  \institution{University of Illinois at Chicago}
  \city{Chicago}
  \state{Illinois}
  \country{USA}
  \postcode{60607}
}

\author{Jaewoo Kang}
\authornote{Corresponding author}
\email{kangj@korea.ac.kr}
\affiliation{%
  \institution{Korea University}
  \city{Seoul}
  \country{South Korea}
  \postcode{02841}
}
\renewcommand{\shortauthors}{Lee et al.}

\begin{abstract}
In the evolving landscape of machine learning, the adaptation of pre-trained models through prompt tuning has become increasingly prominent. 
This trend is particularly observable in the graph domain, where diverse pre-training strategies present unique challenges in developing effective prompt-based tuning methods for graph neural networks. 
Previous approaches have been limited, focusing on specialized prompting functions tailored to models with edge prediction pre-training tasks. 
These methods, however, suffer from a lack of generalizability across different pre-training strategies.
Recently, a simple prompt tuning method has been designed for any pre-training strategy, functioning within the input graph's feature space.
This allows it to theoretically emulate any type of prompting function, thereby significantly increasing its versatility for a range of downstream applications.
Nevertheless, the capacity of such simple prompts to fully grasp the complex contexts found in graphs remains an open question, necessitating further investigation.
Addressing this challenge, our work introduces the Subgraph-level Universal Prompt Tuning (SUPT) approach, focusing on the detailed context within subgraphs.
In SUPT, prompt features are assigned at the subgraph-level, preserving the method's universal capability.
This requires extremely fewer tuning parameters than fine-tuning-based methods, outperforming them in 42 out of 45 full-shot scenario experiments with an average improvement of over 2.5\%.
In few-shot scenarios, it excels in 41 out of 45 experiments, achieving an average performance increase of more than 6.6\%.

\end{abstract}





\maketitle

\newcommand{\checkvalue}[2]{\ifdim #1pt > #2pt \textcolor{blue}{#1}\else \textcolor{red}{#1}\fi}

\section{Introduction}
\label{sec:intro}
Pre-trained models have significantly revolutionized the fields of Natural Language Processing (NLP) and Computer Vision, primarily addressing two critical challenges: the scarcity of labeled data and the need for strong out-of-distribution generalization ability. 
In NLP, models like BERT and GPT have demonstrated remarkable proficiency in understanding and generating human language by leveraging vast, unlabeled datasets \cite{devlin2018bert,brown2020language}. 
Similarly, in Computer Vision, models like ResNet and EfficientNet have shown exceptional performance in image recognition tasks by training on large-scale, diverse image datasets \cite{he2016deep,tan2019efficientnet}. 
Inspired by these successes, the concept of pre-trained models has recently gained traction in the domain of graph representation learning \cite{xia2022survey}. 
The adoption of pre-trained models in this area is particularly promising, considering the complex and often sparse nature of graph data. 
By pre-training on extensive graph datasets, these models can capture intricate patterns and relationships inherent in graph structures, thereby enhancing their performance on various downstream tasks in the graph domain.

\begin{figure}[t]
\begin{center}
\centerline{\includegraphics[width=\columnwidth]{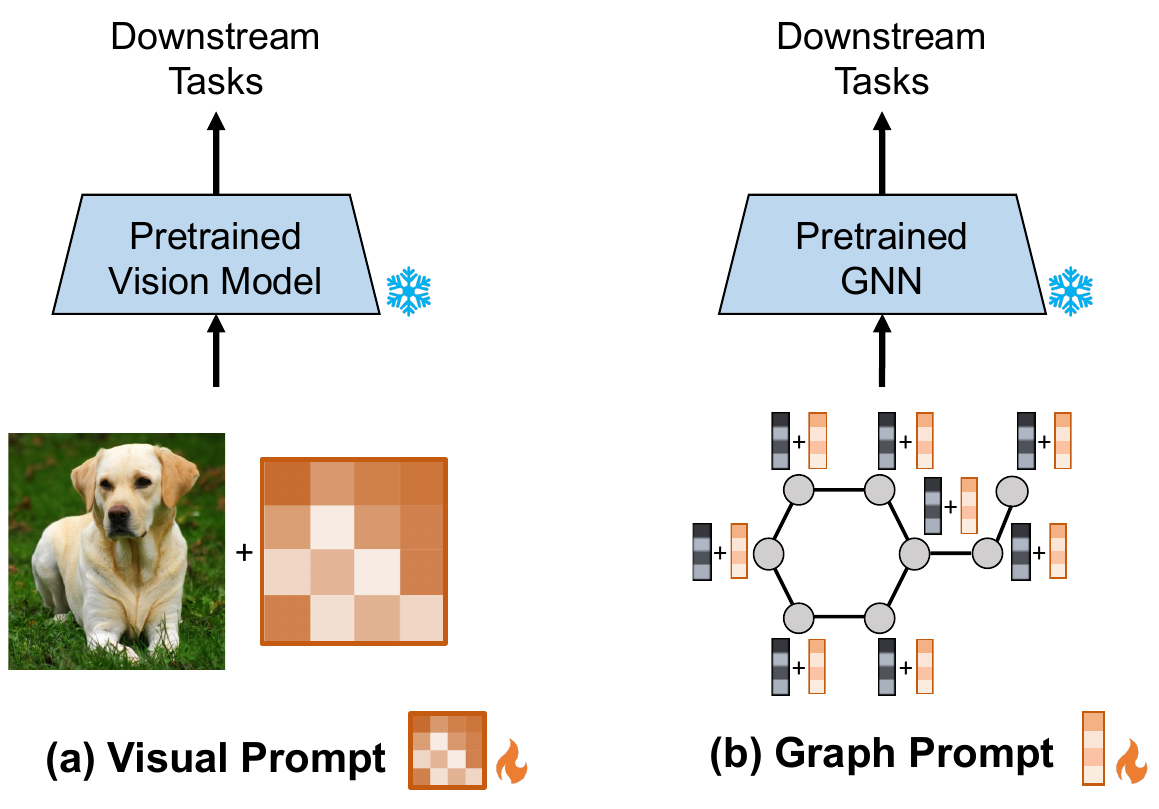}}
\caption{
Illustration of the concepts: (a) pixel-level visual prompts \cite{bahng2022visual,10171397,10097245,liu2023explicit,li2023exploring,Oh_2023_CVPR,tsao2024autovp} and (b) node-level graph prompts \cite{fang2023universal}.
}
\label{fig:VP}
\end{center}
\vskip -0.2in
\end{figure}

\begin{figure*}[t]
\vskip 0.15in
\begin{center}
\centerline{\includegraphics[width=\textwidth]{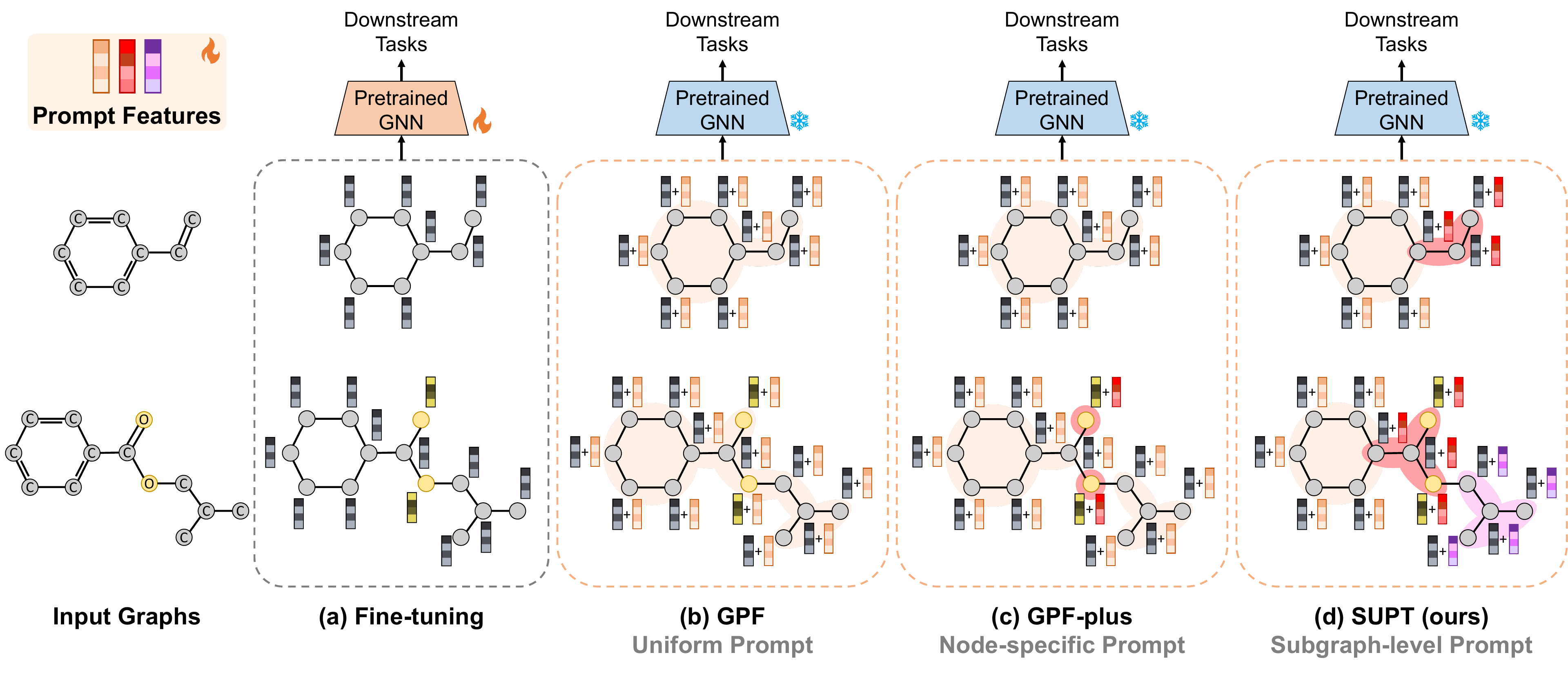}}
\caption{
Illustration of Different Tuning Approaches: Comparing Fine-Tuning, GPF, GPF-Plus, and SUPT.
GPF assigns a uniform prompt feature across all nodes, irrespective of their contextual disparities. 
GPF-Plus advances by allocating prompt features according to node type with category-specific uniformity but not individual graph contextuality.
In contrast, SUPT innovates by assigning prompt features at the subgraph level, capturing the intricate contextual nuances of each subgraph.
This distinction allows SUPT to apply varied prompt features to nodes of the same type, contingent on their subgraph context, thus introducing a layer of context-specific differentiation even among nodes sharing the same type.
}
\label{fig:SUPT}
\end{center}
\vskip -0.1in
\end{figure*}

In recent years, the primary method to leverage pre-trained models in various domains, including graphs, has been the "pre-train, fine-tune" framework \cite{hu2020pretraining}. 
This approach entails initial training of the model on a large and diverse dataset to capture general features, which is subsequently followed by fine-tuning on a smaller, task-specific dataset. 
Although this methodology has demonstrated success, it is not devoid of critical issues.
One of the most significant challenges is the misalignment between the objectives of pre-training tasks and those of downstream tasks. 
For instance, a model pre-trained on generic graph structures may struggle to adapt to specific graph classification tasks due to differing objectives. 
Furthermore, ensuring that the model retains its generalization ability during the fine-tuning process is a non-trivial task. 
As models are fine-tuned on specific datasets, they risk losing their ability to generalize to new, unseen data, a phenomenon known as catastrophic forgetting \cite{liu2021overcoming,zhou2021overcoming}.

To address these challenges, the concept of prompt tuning, "pre-train, prompt, fine-tune", has emerged as a promising solution \cite{li2021prefix,lester2021powerr,liu2023pre}.
Inspired by the success of prompt tuning in NLP, recent studies in computer vision are actively exploring the adaptation of these methods to incorporate visual prompts at the pixel-level, as depicted in Figure \ref{fig:VP} \cite{bahng2022visual,10171397,10097245,liu2023explicit,li2023exploring,Oh_2023_CVPR,tsao2024autovp}.
Unlike the "pre-train, fine-tune" approach, prompt tuning involves modifying the input data to better align with the downstream tasks while keeping the pre-trained model parameters fixed. 
This method has shown potential in maintaining the generalization abilities of the model and ensuring better alignment with specific task objectives.

However, when it comes to Graph Neural Networks (GNNs), the landscape of prompt tuning methods remains largely unexplored, particularly in terms of universal applicability \cite{fang2023universal}. 
Existing prompt tuning methods in GNNs are often tightly coupled with specific pre-training strategies, limiting their broader application \cite{sun2022gppt,liu2023graphprompt}. 
For example, a prompt tuning method designed for a GNN pre-trained on node classification tasks may not be effective for a GNN pre-trained on link prediction tasks. 
This lack of \textit{universal prompt tuning} methods for GNNs stems from the diverse and complex nature of graph data and the corresponding variety of pre-training strategies employed. 
Each graph dataset and task can have unique characteristics and requirements, making the development of a one-size-fits-all prompt tuning approach particularly challenging. 
As a result, there is a pressing need for innovative methods in prompt tuning that can adapt to the wide range of pre-training strategies employed in graph neural networks, thereby unlocking their full potential across various graph-based applications.
Recently, \citeauthor{fang2023universal} introduced the concepts of Graph Prompt Feature (GPF) and its extension (GPF-plus), designed to be theoretically equivalent to any prompting function, thus enhancing their adaptability across various downstream tasks. 
However, the effectiveness of these relatively straightforward prompting mechanisms in capturing the intricate contexts prevalent within graph structures is yet to be fully determined. 
Further investigation is required in this domain, given that GPF utilizes uniform prompting and GPF-plus adopts a node-specific approach, as illustrated in Figure \ref{fig:SUPT}.

In this study, we refine the concept of universal graph prompt tuning by introducing an approach that emphasizes intricacy at the subgraph-level, which we term Subgraph-level Universal Prompt Tuning (SUPT). 
This advancement builds upon existing frameworks like GPF but distinguishes itself by applying prompts at the subgraph-level, enabling more refined control over input features.
Such granularity improves the method's flexibility and utility across a variety of graph-based applications, addressing the complex and diverse nature of graph data.
SUPT not only delivers improved performance across a variety of downstream tasks but also does so while maintaining a small number of parameters comparable to the GPF and GPF-plus. 
Through an extensive evaluation involving 45 experiments across nine datasets and five pre-training methodologies, SUPT stands out by requiring far fewer tuning parameters compared to fine-tuning approaches. 
In full-shot scenarios, it surpasses these conventional methods in 42 out of 45 cases, registering an average performance gain of over 2.5\%. 
The efficacy of SUPT extends into few-shot scenarios as well, where it excels in 41 out of 45 experiments, marking an average performance increment of more than 6.6\%. 
The implementation is available at: https://anonymous.4open.science/r/SUPT-F7B1/

\section{Related works}
\label{sec:related}
\paragraph{\textbf{Graph Pre-training.}}
Influenced by advancements in NLP \cite{qiu2020pre} and Computer Vision \cite{du2022survey}, the graph domain increasingly adopts pre-trained models \cite{xia2022survey}. 
These models, trained on extensive graph datasets, learn to generalize features across a broad range of applications. Various methods for graph pre-training have emerged, showing substantial progress. 
Early efforts included GAE \cite{thomas2016variational} that introduces a method to train graphs using edge prediction as a training objective utilizing variational auto-encoder. 
Deep Graph Infomax \cite{velivckovic2018deep} and InfoGraph \cite{sun2019infograph} focus on maximizing mutual information between different levels of representations to obtain pre-trained graph model. 
Subsequent studies \cite{hu2020gpt, you2020graph} diversify the pre-training tasks to self-supervised attributed graph generation task \cite{hu2020gpt} and contrastive learning with diverse graph augmentation techniques \cite{you2020graph}. 
\citeauthor{hu2020pretraining} propose an effective strategy to pre-train through attribute masking and context prediction, followed by fine-tuning for classification tasks. 
Follow-up studies confirm the effectiveness of "pre-train, fine-tune" framework in adapting pre-trained models to individual downstream tasks \cite{han2021adaptive, xia2022towards}. 
AUX-TS \cite{han2021adaptive} employs a meta-learning based weighting model for adaptive auxiliary task selection during fine-tuning. \citeauthor{xia2022towards} introduce data augmentation and regularization methods to optimize the fine-tuning of pre-trained molecular graphs. 
Nevertheless, challenges arise with the strategy, particularly in aligning various pre-training objectives with specific downstream tasks, which often results in catastrophic forgetting and limited task-specific adaptability \cite{liu2021overcoming}. 
Additionally, as the number of pre-trained parameters increases, fine-tuning for individual downstream tasks becomes more resource-intensive \cite{liu2023pre}.

\paragraph{\textbf{Graph Prompt Tuning.}}
Prompt tuning \cite{liu2023pre}, initially conceptualized in NLP, has been widely applied to enhance the adaptability of pre-trained models to downstream tasks. 
The primary advantage of the method lies in bridging the gap between pre-training and downstream task objectives, requiring fewer parameter updates compared to traditional fine-tuning \cite{liu2023pre}. 
Early approaches utilize hard prompts \cite{brown2020language}, which are in the form of discrete natural language phrases. 
Subsequent research explores soft prompts \cite{lester2021powerr,liu2023gpt,li2021prefix}, employing continuous vectors to surpass the limitations of natural language embeddings. In graph prompt tuning, studies have actively investigated the advantages \cite{sun2022gppt,liu2023graphprompt,fang2023universal}, with a focus on soft prompts due to the unique nature of graph data. 
GPPT \cite{sun2022gppt} introduces a "pre-train, prompt, fine-tune" framework to mitigate the objective gap using soft prompt, but it only deals with link prediction for pre-training and node classification for downstream task. 
GraphPrompt \cite{liu2023graphprompt} expands its application to node and graph classifications, but remains restricted in pre-training objectives. 
\citeauthor{10.1145/3580305.3599256} proposes a multi-task prompting framework that bridges the gap between pre-training and multiple downstream tasks inserting learnable prompt graph into the original graph. 
PRODIGY introduces a framework that introduces in-context learning on graph, making the model to be pre-trained to solve tasks across a wide range of tasks with prompt graph \cite{huang2023prodigy}. 
GPF and GPF-plus \cite{fang2023universal} add soft prompts to all node features of the input graph, enabling the method to be applied to any pre-training strategy. 
Despite these advancements, most methods face limitations in handling diverse pre-training strategies and downstream tasks. 
\citeauthor{fang2023universal} effectively expand the scope of application but face difficulties in embedding unique graph context, restricting soft prompts from capturing context-specific details.

\section{Method}
\label{sec:method}
In this section, we introduce an advanced method for universal graph prompt tuning that is applied to the features of input nodes at the subgraph-level, referred to as Subgraph-level Universal Prompt Tuning (SUPT).
Inspired by the concept of pixel-level prompts for images \cite{bahng2022visual,10171397,10097245,liu2023explicit,li2023exploring,Oh_2023_CVPR,tsao2024autovp} and node-level prompts for graphs \cite{fang2023universal}, SUPT similarly utilizes additional prompt features into the input space of graphs.

\subsection{Preliminaries}
\label{sec:preliminaries}
Graph-structured data can be represented as $\mathcal{G} = (\mathcal{V}, \mathcal{E})$, where $\mathcal{V}$ denotes the set of nodes and $\mathcal{E} \subseteq \mathcal{V} \times \mathcal{V}$ represents the set of edges, respectively. 
The graph includes a node feature matrix $X \in \mathbb{R}^{|\mathcal{V}| \times d}$ and an adjacency matrix $A\in \{0,1\}^{|\mathcal{V}| \times |\mathcal{V}|}$ with feature dimension $d$.
Denoted as $f_\theta$, the pre-trained model is optimized according to the pre-training method's objective with a parameter set $\theta$.

\paragraph{\textbf{Fine-Tuning.}}
In general, fine-tuning methods optimize the parameters of the pre-trained model $f_\theta$ along with a projection head $h_\phi$ parameterized by $\phi$, using a downstream task dataset $\mathcal{D}$ that consists of tuples of graphs and labels $(\mathcal{G}, y)$.
The optimization objective is to maximize the likelihood of correctly predicting labels for graphs in the downstream task, with parameters $\theta$ and $\phi$:
\begin{equation}
    \max_{\theta, \phi} P_{f_\theta,h_\phi}(y|\mathcal{G})
    \label{eq:obj_fine}
\end{equation}

\paragraph{\textbf{Prompt Tuning.}}
In prompt tuning, the input graph $\mathcal{G}$ is modified to prompted graph $g_\psi(\mathcal{G})$ through a parameterized prompt function $g_\psi$, instead of updating the pre-trained model $f_\theta$.
The objective of prompt tuning is to find an optimal task-specific prompted graph $g_\psi(\mathcal{G})$ by maximizing the likelihood with parameters $\psi$ and $\phi$:
\begin{equation}
    \max_{\psi, \phi} P_{f_\theta,h_\phi}(y|g_\psi(\mathcal{G}))
    \label{eq:obj_prompt}
\end{equation}
Note that the number of parameters $\psi$ is significantly smaller than the number of parameters $\theta$ in the pre-trained model.
This is demonstrated in Table \ref{tab:overall}, highlighting the parameter efficiency in the training process of prompt tuning methods compared to fine-tuning methods.

\begin{table*}[h]
    \centering
    \vskip 0.15in
    \begin{tabular}{l c cc c cc c cc}
        \toprule
        &\multirow{2}{*}{\#Params} & \multicolumn{2}{c}{\#Params Ratio(\%)}&& \multicolumn{2}{c}{Full Shot} && \multicolumn{2}{c}{50-Shot} \\
        \cline{3-4} \cline{6-7} \cline{9-10}
        & & Chem & Bio && Outperformance & Gain && Outperformance & Gain \\
        \midrule
        FT& $\sim$ 1.8M-2.7M & 100 & 100 & & - & 0 && - & 0\\
        \midrule
        GPF&  $\sim$ 0.3K & 0.02 & 0.01 && 27 out of 45 & +0.91 && 33 out of 45 & +1.89\\
        GPF-plus & $\sim$ 3-12K & 0.17-0.68 & 0.11-0.44 && 27 out of 45 & +1.06 & & 35 out of 45 & +2.00\\
        \midrule
        SUPT$_{\text{hard}}$ &$\sim$ 0.6-3K & 0.03-0.17 & 0.02-0.11 & & 42 out of 45 & +1.85 && 41 out of 45 & +3.73 \\
        SUPT$_{\text{soft}}$ &$\sim$ 0.6-3K & 0.03-0.17 & 0.02-0.11 & & 43 out of 45 & +2.05 && 41 out of 45 & +3.67 \\
        \bottomrule
    \end{tabular}
    \vskip 0.1in
    \caption{In an overall comparison of the fine-tuning (FT), GPF, and SUPT approaches, it is observed that while FT involves adjusting the model's entire parameter set, prompt tuning strategies like GPF and SUPT employ far fewer additional training parameters. 
    In particular, SUPT stands out by maintaining a small number of prompt parameters yet consistently delivers improved performance, outperforming FT across most experiments.}
    \label{tab:overall}
    \vskip -0.1in
\end{table*}


\paragraph{\textbf{Graph Prompt Feature.}}
Recently, \citeauthor{fang2023universal} introduced a parameterized prompt function known as GPF and GPF-plus.
GPF-plus incorporates the learnable prompt feature vector $p \in \mathbb{R}^d$ into the node feature $x \in \mathbb{R}^d$ as:
\begin{equation}
    X = \{x_1,...,x_N\}, \quad X^* = \{x_1+p_1,...,x_N+p_N\}
    \label{eq:GPF-plus}
\end{equation}
where $N$ is the number of nodes and $p$ is calculated as weighted sum of learnable basis vectors.
The weights for the attentive aggregation of the basis vector $b$ are determined by the softmax values of the learnable linear projection $a$ applied to the node feature $x$.
This can be formulated as: 
\begin{equation}
    p_i=\sum_j^k \alpha_{i,j}b_j , \quad \alpha_{i,j} = \frac{\exp(a_j^\top x_i)}{\sum_l^k \exp(a_l^\top x_i)}
\end{equation}
where $p_i$ denotes the final prompt feature for the $i$-th node, $\alpha_{i,j}$ represents the attention weight of the $j$-th basis for the $i$-th node, and the hyperparameter $k$ is the number of basis vectors and projections.
The weight $\alpha_{i,j}$ varies according to the type of node feature, that is, $x_i$, and therefore, the prompt feature $p_i$ becomes a category-specific prompt for the node.
GPF, applying a uniform prompt feature to all nodes, can be considered a special case of GPF-plus with $k=1$.

\subsection{Subgraph-level Universal Prompt Tuning}
In subgraph-level prompt tuning, we adhere to Equation (\ref{eq:GPF-plus}), which operates in the input space. 
However, instead of utilizing the node feature $x$, we allocate prompt feature vectors based on scores derived from simple GNNs, such as GCN \cite{kipf2017semisupervised} or SGC \cite{wu2019simplifying} for $m$-hop context.
Consequently, the prompt calculation is as follows:
\begin{equation}
    p_i=\sum_j^k \sigma(\alpha)_{i,j}b_j , \quad \alpha = (\tilde{D}^{-\frac{1}{2}}\tilde{A}\tilde{D}^{-\frac{1}{2}})^m(X \oplus \sum_j^k b_j) W 
    \label{eq:SUPT}
\end{equation}
where $\tilde{A}$ and $\tilde{D}$ denote the adjacency matrix and degree matrix with self-loops, respectively, $\oplus$ represents a broadcasted addition, $W \in \mathbb{R}^{d \times k}$ is a weight matrix, $k$ is a hyperparameter defining the number of subgraphs, and $\sigma$ indicates a non-linear operation such as a softmax or an activation function with top-rank selection.
In this context, $W$ and $b$ are learnable parameters of SUPT and correspond to elements of $\psi$ in Equation (\ref{eq:obj_prompt}).
Based on the definition of $\sigma$, we have developed two variants: SUPT$_{\text{soft}}$, which utilizes a softmax function, and SUPT$_{\text{hard}}$, which employs a top-rank selection mechanism.

\paragraph{\textbf{SUPT$_{\text{soft}}$}}
Inspired by cluster-based graph pooling methods \cite{ying2018hierarchical,ma2019graph,bianchi2020spectral,Yuan2020StructPool,Khasahmadi2020MemoryBased}, we view the basis feature $b$ as a prompt assigned to a cluster, which constitutes a subset of nodes.
The non-linear operation $\sigma$ in Equation (\ref{eq:SUPT}) is defined as a column-wise softmax function that is defined as a column-wise softmax function, which ensures that each node is associated with a specific basis.
This setup allows $\alpha \in \mathbb{R}^{N \times k}$ to function similarly to the assignment matrix in soft-cluster graph pooling methods.
Additional techniques, such as entropy regularization \cite{ying2018hierarchical}, can be employed to induce sparsity in $\alpha$, making the basis assignment for each node (e.g. $\alpha_{i,:}$ for $i$-th node) approximate a one-hot vector.
However, in this work, we achieved a sparse implementation by leveraging node-selection graph pooling methods as described below.

\begin{table*}[h]
    \centering
    \vskip 0.15in
    \begin{tabular}{l|ccccccccc}
        \toprule
         Dataset & PPI & BBBP & Tox21 & ToxCast & SIDER & ClinTox & MUV & HIV & BACE  \\
        \midrule 
        \# Proteins / Molecules & 88K & 2039 & 7831 & 8575 & 1427 & 1478 & 93087 & 41127 & 1513  \\
        \# Binary prediction tasks & 40 & 1 & 12 & 617 & 27 & 2 & 17 & 1 & 1  \\
        \bottomrule
    \end{tabular}
    \vskip 0.1in
    \caption{Statistics of downstream task datasets.}
    \label{tab:datasets}
    \vskip -0.1in
\end{table*}

\paragraph{\textbf{SUPT$_{\text{hard}}$}}
We exploit node-selection graph pooling methods \cite{gao2019graph,lee2019self,zhang2020structure,ma2020path} to implement the sparse version of SUPT.
Unlike SUPT$_{\text{soft}}$, SUPT$_{\text{hard}}$ determines the nodes to be assigned to each basis through top-rank selection. 
By performing a row-wise top-rank selection on $\alpha$, we obtain the indices of the subset of nodes corresponding to each basis. 
Nodes that do not correspond to a particular basis do not receive any addition of the prompt.
For instance, considering the $k$-th basis:
\begin{align}
    \text{idx} &= \operatorname{top-rank}(\alpha_{:,k},\lceil rN \rceil) \nonumber \\
    X^*_{\text{idx},:} &= X_{\text{idx},:} + \delta(\alpha_{\text{idx},k}) \odot b_k 
    \label{eq:SUPT_hard}
\end{align}
where $r$, a hyperparameter in the range $(0,1]$, specifies the pooling ratio, $\delta$ represents an activation function such as sigmoid, and $\odot$ denotes element-wise multiplication with broadcasting.
The $\operatorname{top-rank}$ function selects the indices of the top $\lceil rN \rceil$ values in $\alpha_{:,k}\in \mathbb{R}^N$, choosing the nodes for the $k$-th basis.
In SUPT$_{\text{hard}}$, it is possible for some nodes to not be selected by any basis, whereas others could be influenced by prompts from all bases.
To address these disparities, an additional normalization term, such as division by the number of corresponding bases, can be incorporated.

\paragraph{\textbf{Methodological Characteristics of Two Variants}}
The primary distinction between the two variants, SUPT$_{\text{soft}}$ and SUPT$_{\text{hard}}$, lies in their respective approaches to subgraph definition.

SUPT$_{\text{soft}}$ employs a learning approach akin to defining a subset of nodes belonging to the same cluster when determining which nodes will be influenced by the same prompt basis, essentially training an assignment matrix. 
Consequently, every node is influenced by at least one prompt basis, with the extent of influence from any given cluster determined by a softmax function. 
While similar to GPF-plus in that a prompt is added to all nodes, SUPT$_{\text{soft}}$ differentiates itself by determining the assignment score for each basis through a GNN, rather than simply relying on the input node features $x$.

Conversely, SUPT$_{\text{hard}}$ exclusively selects only the top-ranked nodes when determining which ones will be influenced by the same prompt basis. 
Nodes not selected by any basis are input into the pre-trained model without any added prompt, maintaining their original input features. 
This constitutes the technical distinction from GPF, GPF-plus, and SUPT$_{\text{soft}}$.

Both methodologies are inspired by graph pooling methods, yet they do not inherit the associated limitations. 
For instance, cluster-based graph pooling, while beneficial for aggregating node features to create new features, has faced criticism for leading to dense output graphs. 
SUPT$_{\text{soft}}$, on the other hand, merely adds an extra prompt but preserves the sparsity of the graph structure, addressing this concern.
Similarly, node selection-based pooling methods have been criticized for the potential loss of information from nodes that are not selected.
However, this concern does not apply to SUPT$_{\text{hard}}$, which merely refrains from adding prompts to unselected nodes and preserves their original features.
In summary, SUPT is capable of integrating subgraph-level information beneficial for downstream tasks, leveraging an advantage of graph pooling methods without inheriting their significant limitations.

\paragraph{\textbf{Theoretical Analysis}}
In this part, we present a theoretical analysis for SUPT, establishing its universality.
We theoretically show that our methodologies are capable of attaining outcomes equivalent to any prompt function.
This establishes the broad applicability and versatility of our techniques, making them suitable for a variety of pre-training strategies.
Unlike GPF, a notable feature of SUPT is that the prompt is applied only to a subgraph (i.e., a subset of nodes $\mathcal{S} \subseteq \mathcal{V}$), rather than to the entire node set of a given graph. 
A key aspect of our work is to theoretically demonstrate that SUPT maintains universality under these conditions. 
Consequently, this implies that SUPT can also achieve the theoretical performance upper bound for any prompting function, as shown in the GPF paper \cite{fang2023universal}.
In the context of a specified pre-training task $ t $ from a set of tasks $ \mathbb{T} $, and given an input graph $ \mathcal{G} $ characterized by its nodes $ X $ and adjacency matrix $ A $, we introduce a prompting function $ \Psi_t(\cdot) $. 
This function generates a graph template $ \mathcal{G}^* = (X^*, A^*) $, where $ X^* $ is an element of the candidate space $ \mathbb{X} $, and $ A^* $ is an element of the candidate space $ \mathbb{A} $, with $ \mathcal{G}^* = \Psi_t(\mathcal{G}) $.

\begin{theorem}[\textbf{Universal Capability of SUPT}]
Let $ f $ be a pre-trained GNN model and $ \mathcal{G} : (X,A) $ be an input graph with sets of nodes and edges, $\mathcal{V}$ and $\mathcal{E}$. 
Given any prompting function $ \psi_t(\cdot) $, if a prompted graph $ \hat{\mathcal{G}} : (\hat{X} \in \mathbb{X}, \hat{A} \in \mathbb{A}) $ is in the candidate space $ \mathcal{G}^* = \Psi_t(\mathcal{G}) $, then there exists a vector $ \hat{p} $ such that for any non-empty subset of nodes $ \mathcal{S} $ (i.e., $ \mathcal{S} \subseteq \mathcal{V} $ and $ |\mathcal{S}| \geq 1 $), the updated features can be defined as:
\[ 
X' = 
\begin{cases} 
X_s + \hat{p} & \text{if } s \in \mathcal{S} \\
X_s & \text{if } s \notin \mathcal{S} 
\end{cases}
\]
such that there exists $X' $ satisfying:
\[
f(X',A) = f(\hat{X},\hat{A})
\]
\label{th:universality}
\end{theorem}
The proof of Theorem \ref{th:universality} is detailed in the appendix. 
This theorem establishes that applying a prompt to a subgraph can meet the theoretical performance upper bound for any prompting function, as delineated in Equation (\ref{eq:obj_prompt}). 
Essentially, if the optimization of a graph template $\mathcal{G}^*$, formulated by a given prompting function $\Psi(\cdot)$, leads to effective graph representations, then theoretically, the optimization of the prompt vector $p$ within the SUPT framework can also achieve exact graph representations.

\section{Experiments}
\label{sec:exp}

\subsection{Experiment Setup}
For a fair comparison, we strictly adhere to the experimental setup outlined in the GPF paper \cite{fang2023universal}, which is itself based on the implementation developed by \citeauthor{hu2020pretraining}.

\paragraph{\textbf{Datasets}}
In the pre-training phase, \citeauthor{fang2023universal} and \citeauthor{hu2020pretraining} leveraged datasets from both the biological and chemical domains, referred to as the Biology and Chemistry datasets, respectively. 
The Biology dataset is composed of approximately 395k unlabeled protein ego-networks, derived from Protein-Protein Interaction (PPI) networks across 50 distinct species. 
These networks were employed for node-level self-supervised pre-training. 
Additionally, the dataset includes 88k labeled protein ego-networks, utilized for the prediction of 5,000 biological functions, serving the purpose of graph-level multi-task supervised pre-training.
On the chemical side, the Chemistry dataset incorporates around 2 million unlabeled molecules, which were sampled from the ZINC15 database \cite{sterling2015zinc} for the purpose of node-level self-supervised pre-training. 
For graph-level multi-task supervised pre-training, a preprocessed ChEMBL dataset \cite{gaulton2012chembl,mayr2018large} was used. 
This subset comprises 456k molecules with 1,310 biochemical assays, providing a diverse chemical space for comprehensive model training.

For downstream tasks, they utilized the PPI networks dataset, which consists of 88K proteins from 8 different species, to predict 40 biological functions \cite{zitnik2019evolution}, as well as 8 binary classification datasets for molecular property prediction \cite{wu2018moleculenet}.
The statistics of datasets for downstream tasks are summarized in Table \ref{tab:datasets}.

\newcommand{\InfomaxFTBBBPFull}{67.55}
\newcommand{\InfomaxFTToxFull}{78.57}
\newcommand{\InfomaxFTToxCastFull}{65.16}
\newcommand{\InfomaxFTSIDERFull}{63.34}
\newcommand{\InfomaxFTClinToxFull}{70.06}
\newcommand{\InfomaxFTMUVFull}{81.42}
\newcommand{\InfomaxFTHIVFull}{77.71}
\newcommand{\InfomaxFTBACEFull}{81.32}
\newcommand{\InfomaxFTPPIFull}{71.29}

\newcommand{\AttrMaskingFTBBBPFull}{66.33}
\newcommand{\AttrMaskingFTToxFull}{78.28}
\newcommand{\AttrMaskingFTToxCastFull}{65.34}
\newcommand{\AttrMaskingFTSIDERFull}{66.77}
\newcommand{\AttrMaskingFTClinToxFull}{74.46}
\newcommand{\AttrMaskingFTMUVFull}{81.78}
\newcommand{\AttrMaskingFTHIVFull}{77.90}
\newcommand{\AttrMaskingFTBACEFull}{80.94}
\newcommand{\AttrMaskingFTPPIFull}{73.93}

\newcommand{\ContextPredFTBBBPFull}{69.65}
\newcommand{\ContextPredFTToxFull}{78.29}
\newcommand{\ContextPredFTToxCastFull}{66.39}
\newcommand{\ContextPredFTSIDERFull}{64.45}
\newcommand{\ContextPredFTClinToxFull}{73.71}
\newcommand{\ContextPredFTMUVFull}{82.36}
\newcommand{\ContextPredFTHIVFull}{79.20}
\newcommand{\ContextPredFTBACEFull}{84.66}
\newcommand{\ContextPredFTPPIFull}{72.10}

\newcommand{\GCLFTBBBPFull}{69.49}
\newcommand{\GCLFTToxFull}{73.35}
\newcommand{\GCLFTToxCastFull}{62.54}
\newcommand{\GCLFTSIDERFull}{60.63}
\newcommand{\GCLFTClinToxFull}{75.17}
\newcommand{\GCLFTMUVFull}{69.78}
\newcommand{\GCLFTHIVFull}{78.26}
\newcommand{\GCLFTBACEFull}{75.51}
\newcommand{\GCLFTPPIFull}{67.76}

\newcommand{\EdgePredFTBBBPFull}{66.56}
\newcommand{\EdgePredFTToxFull}{78.67}
\newcommand{\EdgePredFTToxCastFull}{66.29}
\newcommand{\EdgePredFTSIDERFull}{64.35}
\newcommand{\EdgePredFTClinToxFull}{69.07}
\newcommand{\EdgePredFTMUVFull}{79.67}
\newcommand{\EdgePredFTHIVFull}{77.44}
\newcommand{\EdgePredFTBACEFull}{80.90}
\newcommand{\EdgePredFTPPIFull}{71.54}

\begin{table*}[t]
    \centering
    \vskip 0.15in
    \begin{tabular}{ll|ccccccccc|r}
        \toprule
         & Tuning & PPI & BBBP & Tox21 & ToxCast & SIDER & ClinTox & MUV & HIV & BACE & Avg. \\
        \midrule 
        
        \multirow{5}{*}{\rotatebox{90}{Infomax}} & FT & \InfomaxFTPPIFull & \InfomaxFTBBBPFull & \InfomaxFTToxFull & \InfomaxFTToxCastFull & \InfomaxFTSIDERFull & \InfomaxFTClinToxFull & \InfomaxFTMUVFull & \InfomaxFTHIVFull & \InfomaxFTBACEFull & 72.94 \\
        & GPF & \checkvalue{78.87}{\InfomaxFTPPIFull} & \checkvalue{66.58}{\InfomaxFTBBBPFull} & \checkvalue{78.13}{\InfomaxFTToxFull} & \checkvalue{65.96}{\InfomaxFTToxCastFull} & \checkvalue{65.64}{\InfomaxFTSIDERFull} & \checkvalue{74.14}{\InfomaxFTClinToxFull} & \checkvalue{80.39}{\InfomaxFTMUVFull} & \checkvalue{76.67}{\InfomaxFTHIVFull} & \checkvalue{83.24}{\InfomaxFTBACEFull} & 74.40 \\
        & GPF-plus & \checkvalue{78.85}{\InfomaxFTPPIFull} & \checkvalue{67.03}{\InfomaxFTBBBPFull} & \checkvalue{78.50}{\InfomaxFTToxFull} & \checkvalue{65.83}{\InfomaxFTToxCastFull} & \checkvalue{65.89}{\InfomaxFTSIDERFull} & \checkvalue{73.61}{\InfomaxFTClinToxFull} & \checkvalue{79.94}{\InfomaxFTMUVFull} & \checkvalue{75.60}{\InfomaxFTHIVFull} & \checkvalue{82.96}{\InfomaxFTBACEFull} & 74.25 \\
        \cline{2-12} 
        & SUPT$_{\text{hard}}$ & \checkvalue{79.64}{\InfomaxFTPPIFull} & \checkvalue{67.29}{\InfomaxFTBBBPFull} & \checkvalue{79.14}{\InfomaxFTToxFull} & \checkvalue{66.48}{\InfomaxFTToxCastFull} & \checkvalue{65.78}{\InfomaxFTSIDERFull} & \checkvalue{75.65}{\InfomaxFTClinToxFull} & \checkvalue{81.51}{\InfomaxFTMUVFull} & \checkvalue{77.96}{\InfomaxFTHIVFull} & \checkvalue{84.05}{\InfomaxFTBACEFull} & 75.28 \\
        & SUPT$_{\text{soft}}$ & \checkvalue{79.45}{\InfomaxFTPPIFull} & \checkvalue{67.63}{\InfomaxFTBBBPFull} & \checkvalue{78.98}{\InfomaxFTToxFull} & \checkvalue{66.53}{\InfomaxFTToxCastFull} & \checkvalue{65.88}{\InfomaxFTSIDERFull} & \checkvalue{76.08}{\InfomaxFTClinToxFull} & \checkvalue{81.59}{\InfomaxFTMUVFull} & \checkvalue{77.94}{\InfomaxFTHIVFull} & \checkvalue{84.01}{\InfomaxFTBACEFull} & 75.34 \\
        \midrule
        
        \multirow{5}{*}{\rotatebox{90}{AttrMasking}} & FT & \AttrMaskingFTPPIFull & \AttrMaskingFTBBBPFull & \AttrMaskingFTToxFull & \AttrMaskingFTToxCastFull & \AttrMaskingFTSIDERFull & \AttrMaskingFTClinToxFull & \AttrMaskingFTMUVFull & \AttrMaskingFTHIVFull & \AttrMaskingFTBACEFull & 73.97 \\
        & GPF & \checkvalue{81.47}{\AttrMaskingFTPPIFull} & \checkvalue{67.49}{\AttrMaskingFTBBBPFull} & \checkvalue{78.66}{\AttrMaskingFTToxFull} & \checkvalue{66.09}{\AttrMaskingFTToxCastFull} & \checkvalue{68.19}{\AttrMaskingFTSIDERFull} & \checkvalue{72.72}{\AttrMaskingFTClinToxFull} & \checkvalue{81.44}{\AttrMaskingFTMUVFull} & \checkvalue{77.09}{\AttrMaskingFTHIVFull} & \checkvalue{84.77}{\AttrMaskingFTBACEFull} & 75.32 \\
        & GPF-plus & \checkvalue{81.43}{\AttrMaskingFTPPIFull} & \checkvalue{67.32}{\AttrMaskingFTBBBPFull} & \checkvalue{78.83}{\AttrMaskingFTToxFull} & \checkvalue{66.53}{\AttrMaskingFTToxCastFull} & \checkvalue{68.53}{\AttrMaskingFTSIDERFull} & \checkvalue{74.17}{\AttrMaskingFTClinToxFull} & \checkvalue{81.27}{\AttrMaskingFTMUVFull} & \checkvalue{77.64}{\AttrMaskingFTHIVFull} & \checkvalue{85.01}{\AttrMaskingFTBACEFull} & 75.64 \\
        \cline{2-12} 
        & SUPT$_{\text{hard}}$ & \checkvalue{81.90}{\AttrMaskingFTPPIFull} & \checkvalue{68.39}{\AttrMaskingFTBBBPFull} & \checkvalue{79.50}{\AttrMaskingFTToxFull} & \checkvalue{66.79}{\AttrMaskingFTToxCastFull} & \checkvalue{68.82}{\AttrMaskingFTSIDERFull} & \checkvalue{75.25}{\AttrMaskingFTClinToxFull} & \checkvalue{81.81}{\AttrMaskingFTMUVFull} & \checkvalue{79.10}{\AttrMaskingFTHIVFull} & \checkvalue{84.98}{\AttrMaskingFTBACEFull} & 76.28\\
        & SUPT$_{\text{soft}}$& \checkvalue{81.91}{\AttrMaskingFTPPIFull} & \checkvalue{68.37}{\AttrMaskingFTBBBPFull} & \checkvalue{79.73}{\AttrMaskingFTToxFull} & \checkvalue{67.02}{\AttrMaskingFTToxCastFull} & \checkvalue{68.68}{\AttrMaskingFTSIDERFull} & \checkvalue{77.36}{\AttrMaskingFTClinToxFull} & \checkvalue{81.93}{\AttrMaskingFTMUVFull} & \checkvalue{79.15}{\AttrMaskingFTHIVFull} & \checkvalue{85.22}{\AttrMaskingFTBACEFull} & 76.60 \\
        \midrule
        
        \multirow{5}{*}{\rotatebox{90}{ContextPred}} & FT & \ContextPredFTPPIFull & \ContextPredFTBBBPFull & \ContextPredFTToxFull & \ContextPredFTToxCastFull & \ContextPredFTSIDERFull & \ContextPredFTClinToxFull & \ContextPredFTMUVFull & \ContextPredFTHIVFull & \ContextPredFTBACEFull & 74.53 \\
        & GPF & \checkvalue{80.21}{\ContextPredFTPPIFull} & \checkvalue{69.04}{\ContextPredFTBBBPFull} & \checkvalue{79.90}{\ContextPredFTToxFull} & \checkvalue{67.68}{\ContextPredFTToxCastFull} & \checkvalue{65.31}{\ContextPredFTSIDERFull} & \checkvalue{73.15}{\ContextPredFTClinToxFull} & \checkvalue{84.22}{\ContextPredFTMUVFull} & \checkvalue{77.77}{\ContextPredFTHIVFull} & \checkvalue{85.03}{\ContextPredFTBACEFull} & 75.81 \\
        & GPF-plus & \checkvalue{80.28}{\ContextPredFTPPIFull} & \checkvalue{68.59}{\ContextPredFTBBBPFull} & \checkvalue{79.95}{\ContextPredFTToxFull} & \checkvalue{67.72}{\ContextPredFTToxCastFull} & \checkvalue{66.32}{\ContextPredFTSIDERFull} & \checkvalue{72.73}{\ContextPredFTClinToxFull} & \checkvalue{84.40}{\ContextPredFTMUVFull} & \checkvalue{78.10}{\ContextPredFTHIVFull} & \checkvalue{84.50}{\ContextPredFTBACEFull} & 75.84 \\
        \cline{2-12} 
        & SUPT$_{\text{hard}}$ & \checkvalue{80.85}{\ContextPredFTPPIFull} & \checkvalue{70.18}{\ContextPredFTBBBPFull} & \checkvalue{80.00}{\ContextPredFTToxFull} & \checkvalue{68.40}{\ContextPredFTToxCastFull} & \checkvalue{66.37}{\ContextPredFTSIDERFull} & \checkvalue{74.90}{\ContextPredFTClinToxFull} & \checkvalue{84.54}{\ContextPredFTMUVFull} & \checkvalue{79.42}{\ContextPredFTHIVFull} & \checkvalue{85.21}{\ContextPredFTBACEFull} & 76.65\\
        & SUPT$_{\text{soft}}$ & \checkvalue{80.86}{\ContextPredFTPPIFull} & \checkvalue{70.06}{\ContextPredFTBBBPFull} & \checkvalue{80.12}{\ContextPredFTToxFull} & \checkvalue{68.50}{\ContextPredFTToxCastFull} & \checkvalue{66.48}{\ContextPredFTSIDERFull} & \checkvalue{75.55}{\ContextPredFTClinToxFull} & \checkvalue{84.67}{\ContextPredFTMUVFull} & \checkvalue{79.39}{\ContextPredFTHIVFull} & \checkvalue{85.27}{\ContextPredFTBACEFull} & 76.77\\

        \midrule
        \multirow{5}{*}{\rotatebox{90}{GCL}} & FT & \GCLFTPPIFull & \GCLFTBBBPFull & \GCLFTToxFull & \GCLFTToxCastFull & \GCLFTSIDERFull & \GCLFTClinToxFull & \GCLFTMUVFull & \GCLFTHIVFull & \GCLFTBACEFull & 70.27 \\
        & GPF & \checkvalue{67.91}{\GCLFTPPIFull} & \checkvalue{68.11}{\GCLFTBBBPFull} & \checkvalue{72.15}{\GCLFTToxFull} & \checkvalue{60.73}{\GCLFTToxCastFull} & \checkvalue{61.00}{\GCLFTSIDERFull} & \checkvalue{69.48}{\GCLFTClinToxFull} & \checkvalue{70.81}{\GCLFTMUVFull} & \checkvalue{75.12}{\GCLFTHIVFull} & \checkvalue{76.67}{\GCLFTBACEFull} & 69.11 \\
        & GPF-plus & \checkvalue{66.71}{\GCLFTPPIFull} & \checkvalue{67.77}{\GCLFTBBBPFull} & \checkvalue{73.06}{\GCLFTToxFull} & \checkvalue{61.47}{\GCLFTToxCastFull} & \checkvalue{61.57}{\GCLFTSIDERFull} & \checkvalue{68.89}{\GCLFTClinToxFull} & \checkvalue{70.54}{\GCLFTMUVFull} & \checkvalue{75.90}{\GCLFTHIVFull} & \checkvalue{77.84}{\GCLFTBACEFull} & 69.31 \\
        \cline{2-12} 
        & SUPT$_{\text{hard}}$ & \checkvalue{68.59}{\GCLFTPPIFull} & \checkvalue{69.68}{\GCLFTBBBPFull} & \checkvalue{73.44}{\GCLFTToxFull} & \checkvalue{63.35}{\GCLFTToxCastFull} & \checkvalue{62.51}{\GCLFTSIDERFull} & \checkvalue{69.47}{\GCLFTClinToxFull} & \checkvalue{70.64}{\GCLFTMUVFull} & \checkvalue{76.22}{\GCLFTHIVFull} & \checkvalue{79.17}{\GCLFTBACEFull} & 70.34 \\
        & SUPT$_{\text{soft}}$ & \checkvalue{69.57}{\GCLFTPPIFull} & \checkvalue{69.79}{\GCLFTBBBPFull} & \checkvalue{73.40}{\GCLFTToxFull} & \checkvalue{63.42}{\GCLFTToxCastFull} & \checkvalue{62.84}{\GCLFTSIDERFull} & \checkvalue{71.21}{\GCLFTClinToxFull} & \checkvalue{70.92}{\GCLFTMUVFull} & \checkvalue{76.44}{\GCLFTHIVFull} & \checkvalue{78.92}{\GCLFTBACEFull} & 70.72 \\
        \midrule
        
        \multirow{5}{*}{\rotatebox{90}{EdgePred}} & FT & \EdgePredFTPPIFull & \EdgePredFTBBBPFull & \EdgePredFTToxFull & \EdgePredFTToxCastFull & \EdgePredFTSIDERFull & \EdgePredFTClinToxFull & \EdgePredFTMUVFull & \EdgePredFTHIVFull & \EdgePredFTBACEFull & 72.72 \\
        & GPF & \checkvalue{79.96}{\EdgePredFTPPIFull} & \checkvalue{69.36}{\EdgePredFTBBBPFull} & \checkvalue{79.69}{\EdgePredFTToxFull} & \checkvalue{65.70}{\EdgePredFTToxCastFull} & \checkvalue{66.71}{\EdgePredFTSIDERFull} & \checkvalue{67.92}{\EdgePredFTClinToxFull} & \checkvalue{82.61}{\EdgePredFTMUVFull} & \checkvalue{77.53}{\EdgePredFTHIVFull} & \checkvalue{79.76}{\EdgePredFTBACEFull} & 74.36 \\
        & GPF-plus & \checkvalue{79.88}{\EdgePredFTPPIFull} & \checkvalue{69.31}{\EdgePredFTBBBPFull} & \checkvalue{79.91}{\EdgePredFTToxFull} & \checkvalue{65.66}{\EdgePredFTToxCastFull} & \checkvalue{66.61}{\EdgePredFTSIDERFull} & \checkvalue{69.58}{\EdgePredFTClinToxFull} & \checkvalue{82.74}{\EdgePredFTMUVFull} & \checkvalue{77.77}{\EdgePredFTHIVFull} & \checkvalue{80.91}{\EdgePredFTBACEFull} & 74.71 \\
        \cline{2-12} 
        & SUPT$_{\text{hard}}$ & \checkvalue{80.20}{\EdgePredFTPPIFull} & \checkvalue{69.51}{\EdgePredFTBBBPFull} & \checkvalue{80.76}{\EdgePredFTToxFull} & \checkvalue{66.44}{\EdgePredFTToxCastFull} & \checkvalue{66.85}{\EdgePredFTSIDERFull} & \checkvalue{69.97}{\EdgePredFTClinToxFull} & \checkvalue{82.94}{\EdgePredFTMUVFull} & \checkvalue{77.79}{\EdgePredFTHIVFull} & \checkvalue{81.72}{\EdgePredFTBACEFull} & 75.13 \\
        & SUPT$_{\text{soft}}$ & \checkvalue{80.19}{\EdgePredFTPPIFull} & \checkvalue{69.93}{\EdgePredFTBBBPFull} & \checkvalue{80.90}{\EdgePredFTToxFull} & \checkvalue{66.65}{\EdgePredFTToxCastFull} & \checkvalue{66.95}{\EdgePredFTSIDERFull} & \checkvalue{70.41}{\EdgePredFTClinToxFull} & \checkvalue{82.95}{\EdgePredFTMUVFull} & \checkvalue{77.95}{\EdgePredFTHIVFull} & \checkvalue{81.44}{\EdgePredFTBACEFull} & 75.26 \\
        \bottomrule
    \end{tabular}
    \vskip 0.1in
    \caption{Comparison of test ROC-AUC performance in full-shot scenarios across protein function prediction and molecular prediction benchmarks using various pre-training strategies, highlighted in blue for performances superior to fine-tuning (FT) and in red otherwise. 
    The reported values are the averages obtained from experiments repeated with 20 random seeds. 
    Full results including comparisons with other tuning methods such as GPPT and GraphPrompt, are provided in the Appendix.}
    
    \vskip -0.1in
    \label{tab:full}
\end{table*}

\paragraph{\textbf{Pre-training Methods}}
We employ the model weights (checkpoints) of the pre-trained models provided by \citeauthor{fang2023universal} specifically to assess the effectiveness of various prompt tuning methods, while controlling for other variables.
In their study, the evaluation of GPF across five distinct pre-training strategies was conducted to establish its universal applicability.
The first of these, Deep Graph Infomax (referred to as Infomax), aims to refine node or graph representations by optimizing the mutual information between the representations of entire graphs and their substructures across various levels \cite{velivckovic2018deep}.
Attribute Masking (AttrMasking) involves masking attributes of nodes or edges and tasking the GNN with inferring these attributes using the adjacent structural information \cite{hu2020pretraining}.
Similarly, Context Prediction (ContextPred), also introduced by \citeauthor{hu2020pretraining}, uses subgraphs to forecast their surrounding graph contexts, aiming to align nodes that appear in similar structural contexts closely within the embedding space.
Graph Contrastive Learning (GCL), embeds augmented versions of an anchor graph in close proximity (as positive samples) while distancing the embeddings of negative samples \cite{you2020graph}.
Lastly, Edge Prediction (EdgePred), commonly employed in graph reconstruction studies by the Graph Autoencoder (GAE), focuses on predicting the presence or absence of edges between pairs of nodes \cite{thomas2016variational}. 

\paragraph{\textbf{Model Architecture and Training Scheme}}
Consistent with the approach adopted by \citeauthor{fang2023universal}, our research utilizes the predominant 5-layer GIN architecture \cite{xu2018how}, which is widely acknowledged in pre-trained GNN studies \cite{qiu2020gcc,you2020graph,hu2020pretraining,suresh2021adversarial,xu2021self,zhang2021motif,you2022bringing,xia2022simgrace}. 
While constructing larger GNN models by addressing the challenges such as over-smoothing or over-squashing  is a significant research endeavor in itself \cite{xu2018representation,liu2020towards,zhou2021understanding,li2021training,wu2023demystifying}, the primary focus of this study is to explore more informative prompt tuning methods when models are pre-trained through various strategies.

\newcommand{\InfomaxFTBBBPFew}{53.81}
\newcommand{\InfomaxFTToxFew}{61.42}
\newcommand{\InfomaxFTToxCastFew}{53.93}
\newcommand{\InfomaxFTSIDERFew}{50.77}
\newcommand{\InfomaxFTClinToxFew}{58.60}
\newcommand{\InfomaxFTMUVFew}{66.12}
\newcommand{\InfomaxFTHIVFew}{65.09}
\newcommand{\InfomaxFTBACEFew}{52.64}
\newcommand{\InfomaxFTPPIFew}{48.79}

\newcommand{\AttrMaskingFTBBBPFew}{51.26}
\newcommand{\AttrMaskingFTToxFew}{60.28}
\newcommand{\AttrMaskingFTToxCastFew}{53.47}
\newcommand{\AttrMaskingFTSIDERFew}{50.11}
\newcommand{\AttrMaskingFTClinToxFew}{61.51}
\newcommand{\AttrMaskingFTMUVFew}{59.35}
\newcommand{\AttrMaskingFTHIVFew}{67.18}
\newcommand{\AttrMaskingFTBACEFew}{55.62}
\newcommand{\AttrMaskingFTPPIFew}{48.17}

\newcommand{\ContextPredFTBBBPFew}{49.45}
\newcommand{\ContextPredFTToxFew}{58.77}
\newcommand{\ContextPredFTToxCastFew}{54.46}
\newcommand{\ContextPredFTSIDERFew}{49.89}
\newcommand{\ContextPredFTClinToxFew}{48.60}
\newcommand{\ContextPredFTMUVFew}{56.14}
\newcommand{\ContextPredFTHIVFew}{60.91}
\newcommand{\ContextPredFTBACEFew}{56.37}
\newcommand{\ContextPredFTPPIFew}{46.33}

\newcommand{\GCLFTBBBPFew}{54.40}
\newcommand{\GCLFTToxFew}{48.35}
\newcommand{\GCLFTToxCastFew}{50.29}
\newcommand{\GCLFTSIDERFew}{53.23}
\newcommand{\GCLFTClinToxFew}{54.05}
\newcommand{\GCLFTMUVFew}{46.73}
\newcommand{\GCLFTHIVFew}{60.05}
\newcommand{\GCLFTBACEFew}{49.87}
\newcommand{\GCLFTPPIFew}{49.94}

\newcommand{\EdgePredFTBBBPFew}{48.88}
\newcommand{\EdgePredFTToxFew}{60.95}
\newcommand{\EdgePredFTToxCastFew}{55.73}
\newcommand{\EdgePredFTSIDERFew}{51.30}
\newcommand{\EdgePredFTClinToxFew}{57.78}
\newcommand{\EdgePredFTMUVFew}{66.88}
\newcommand{\EdgePredFTHIVFew}{64.22}
\newcommand{\EdgePredFTBACEFew}{61.27}
\newcommand{\EdgePredFTPPIFew}{47.62}

\begin{table*}[h]
    \centering
    \vskip 0.15in
    \begin{tabular}{ll|ccccccccc|r}
    \toprule
    & Tuning & PPI &BBBP& Tox21 & ToxCast & SIDER & ClinTox & MUV & HIV & BACE & Avg. \\
    \midrule
    
    \multirow{5}{*}{\rotatebox{90}{Infomax}} & FT & \InfomaxFTPPIFew & \InfomaxFTBBBPFew & \InfomaxFTToxFew & \InfomaxFTToxCastFew & \InfomaxFTSIDERFew & \InfomaxFTClinToxFew & \InfomaxFTMUVFew & \InfomaxFTHIVFew & \InfomaxFTBACEFew & 56.80 \\
    & GPF & \checkvalue{52.39}{\InfomaxFTPPIFew} & \checkvalue{56.18}{\InfomaxFTBBBPFew} & \checkvalue{65.42}{\InfomaxFTToxFew} & \checkvalue{56.80}{\InfomaxFTToxCastFew} & \checkvalue{50.19}{\InfomaxFTSIDERFew} & \checkvalue{62.40}{\InfomaxFTClinToxFew} & \checkvalue{68.00}{\InfomaxFTMUVFew} & \checkvalue{66.97}{\InfomaxFTHIVFew} & \checkvalue{51.13}{\InfomaxFTBACEFew} & 58.83 \\
    & GPF-plus & \checkvalue{52.44}{\InfomaxFTPPIFew} & \checkvalue{56.25}{\InfomaxFTBBBPFew} & \checkvalue{65.19}{\InfomaxFTToxFew} & \checkvalue{56.30}{\InfomaxFTToxCastFew} & \checkvalue{50.20}{\InfomaxFTSIDERFew} & \checkvalue{62.34}{\InfomaxFTClinToxFew} & \checkvalue{67.99}{\InfomaxFTMUVFew} & \checkvalue{66.71}{\InfomaxFTHIVFew} & \checkvalue{53.53}{\InfomaxFTBACEFew} & 58.99 \\
    \cline{2-12} 
    & SUPT$_{\text{hard}}$ & \checkvalue{56.16}{\InfomaxFTPPIFew} & \checkvalue{57.12}{\InfomaxFTBBBPFew} & \checkvalue{65.64}{\InfomaxFTToxFew} & \checkvalue{57.16}{\InfomaxFTToxCastFew} & \checkvalue{51.56}{\InfomaxFTSIDERFew} & \checkvalue{64.55}{\InfomaxFTClinToxFew} & \checkvalue{67.97}{\InfomaxFTMUVFew} & \checkvalue{67.03}{\InfomaxFTHIVFew} & \checkvalue{57.03}{\InfomaxFTBACEFew} & 60.47 \\
    & SUPT$_{\text{soft}}$ & \checkvalue{56.29}{\InfomaxFTPPIFew} & \checkvalue{57.06}{\InfomaxFTBBBPFew} & \checkvalue{65.92}{\InfomaxFTToxFew} & \checkvalue{57.16}{\InfomaxFTToxCastFew} & \checkvalue{51.51}{\InfomaxFTSIDERFew} & \checkvalue{64.44}{\InfomaxFTClinToxFew} & \checkvalue{68.27}{\InfomaxFTMUVFew} & \checkvalue{67.04}{\InfomaxFTHIVFew} & \checkvalue{56.59}{\InfomaxFTBACEFew} & 60.48 \\
    \midrule
    
    \multirow{5}{*}{\rotatebox{90}{AttrMasking}} & FT & \AttrMaskingFTPPIFew & \AttrMaskingFTBBBPFew & \AttrMaskingFTToxFew & \AttrMaskingFTToxCastFew & \AttrMaskingFTSIDERFew & \AttrMaskingFTClinToxFew & \AttrMaskingFTMUVFew & \AttrMaskingFTHIVFew & \AttrMaskingFTBACEFew & 56.33 \\
    & GPF & \checkvalue{52.05}{\AttrMaskingFTPPIFew} & \checkvalue{52.60}{\AttrMaskingFTBBBPFew} & \checkvalue{64.35}{\AttrMaskingFTToxFew} & \checkvalue{56.69}{\AttrMaskingFTToxCastFew} & \checkvalue{49.61}{\AttrMaskingFTSIDERFew} & \checkvalue{64.80}{\AttrMaskingFTClinToxFew} & \checkvalue{60.84}{\AttrMaskingFTMUVFew} & \checkvalue{67.74}{\AttrMaskingFTHIVFew} & \checkvalue{53.04}{\AttrMaskingFTBACEFew} &  57.97 \\
    & GPF-plus & \checkvalue{52.12}{\AttrMaskingFTPPIFew} & \checkvalue{53.43}{\AttrMaskingFTBBBPFew} & \checkvalue{64.29}{\AttrMaskingFTToxFew} & \checkvalue{56.71}{\AttrMaskingFTToxCastFew} & \checkvalue{50.19}{\AttrMaskingFTSIDERFew} & \checkvalue{62.92}{\AttrMaskingFTClinToxFew} & \checkvalue{62.10}{\AttrMaskingFTMUVFew} & \checkvalue{68.19}{\AttrMaskingFTHIVFew} & \checkvalue{51.28}{\AttrMaskingFTBACEFew} &  57.91 \\
    \cline{2-12} 
    & SUPT$_{\text{hard}}$ & \checkvalue{56.58}{\AttrMaskingFTPPIFew} & \checkvalue{53.12}{\AttrMaskingFTBBBPFew} & \checkvalue{64.66}{\AttrMaskingFTToxFew} & \checkvalue{56.74}{\AttrMaskingFTToxCastFew} & \checkvalue{51.54}{\AttrMaskingFTSIDERFew} & \checkvalue{66.35}{\AttrMaskingFTClinToxFew} & \checkvalue{62.20}{\AttrMaskingFTMUVFew} & \checkvalue{67.99}{\AttrMaskingFTHIVFew} & \checkvalue{56.57}{\AttrMaskingFTBACEFew} & 59.53 \\
    & SUPT$_{\text{soft}}$ & \checkvalue{56.58}{\AttrMaskingFTPPIFew} & \checkvalue{54.10}{\AttrMaskingFTBBBPFew} & \checkvalue{64.57}{\AttrMaskingFTToxFew} & \checkvalue{56.76}{\AttrMaskingFTToxCastFew} & \checkvalue{51.66}{\AttrMaskingFTSIDERFew} & \checkvalue{66.27}{\AttrMaskingFTClinToxFew} & \checkvalue{62.15}{\AttrMaskingFTMUVFew} & \checkvalue{68.05}{\AttrMaskingFTHIVFew} & \checkvalue{56.80}{\AttrMaskingFTBACEFew} & 59.66 \\
    \midrule
    
    \multirow{5}{*}{\rotatebox{90}{ContextPred}} & FT & \ContextPredFTPPIFew & \ContextPredFTBBBPFew & \ContextPredFTToxFew & \ContextPredFTToxCastFew & \ContextPredFTSIDERFew & \ContextPredFTClinToxFew & \ContextPredFTMUVFew & \ContextPredFTHIVFew & \ContextPredFTBACEFew & 53.44 \\
    & GPF & \checkvalue{51.42}{\ContextPredFTPPIFew} & \checkvalue{50.99}{\ContextPredFTBBBPFew} & \checkvalue{59.76}{\ContextPredFTToxFew} & \checkvalue{55.31}{\ContextPredFTToxCastFew} & \checkvalue{50.23}{\ContextPredFTSIDERFew} & \checkvalue{52.03}{\ContextPredFTClinToxFew} & \checkvalue{60.26}{\ContextPredFTMUVFew} & \checkvalue{59.62}{\ContextPredFTHIVFew} & \checkvalue{56.19}{\ContextPredFTBACEFew} & 55.09 \\
    & GPF-plus & \checkvalue{51.48}{\ContextPredFTPPIFew} & \checkvalue{52.15}{\ContextPredFTBBBPFew} & \checkvalue{59.77}{\ContextPredFTToxFew} & \checkvalue{54.87}{\ContextPredFTToxCastFew} & \checkvalue{50.19}{\ContextPredFTSIDERFew} & \checkvalue{51.71}{\ContextPredFTClinToxFew} & \checkvalue{60.05}{\ContextPredFTMUVFew} & \checkvalue{60.19}{\ContextPredFTHIVFew} & \checkvalue{55.49}{\ContextPredFTBACEFew} & 55.10 \\
    \cline{2-12} 
    & SUPT$_{\text{hard}}$ & \checkvalue{56.12}{\ContextPredFTPPIFew} & \checkvalue{52.29}{\ContextPredFTBBBPFew} & \checkvalue{60.34}{\ContextPredFTToxFew} & \checkvalue{55.46}{\ContextPredFTToxCastFew} & \checkvalue{52.25}{\ContextPredFTSIDERFew} & \checkvalue{57.87}{\ContextPredFTClinToxFew} & \checkvalue{62.54}{\ContextPredFTMUVFew} & \checkvalue{60.64}{\ContextPredFTHIVFew} & \checkvalue{58.97}{\ContextPredFTBACEFew} & 57.39 \\
    & SUPT$_{\text{soft}}$ & \checkvalue{56.15}{\ContextPredFTPPIFew} & \checkvalue{52.68}{\ContextPredFTBBBPFew} & \checkvalue{60.72}{\ContextPredFTToxFew} & \checkvalue{55.39}{\ContextPredFTToxCastFew} & \checkvalue{52.25}{\ContextPredFTSIDERFew} & \checkvalue{53.43}{\ContextPredFTClinToxFew} & \checkvalue{61.32}{\ContextPredFTMUVFew} & \checkvalue{60.82}{\ContextPredFTHIVFew} & \checkvalue{58.31}{\ContextPredFTBACEFew} & 56.79 \\
    \midrule
    
    \multirow{5}{*}{\rotatebox{90}{GCL}}  & FT & \GCLFTPPIFew & \GCLFTBBBPFew & \GCLFTToxFew & \GCLFTToxCastFew & \GCLFTSIDERFew & \GCLFTClinToxFew & \GCLFTMUVFew & \GCLFTHIVFew & \GCLFTBACEFew & 51.88 \\
    & GPF & \checkvalue{46.00}{\GCLFTPPIFew} & \checkvalue{53.61}{\GCLFTBBBPFew} & \checkvalue{50.23}{\GCLFTToxFew} & \checkvalue{53.92}{\GCLFTToxCastFew} & \checkvalue{50.56}{\GCLFTSIDERFew} & \checkvalue{53.21}{\GCLFTClinToxFew} & \checkvalue{51.92}{\GCLFTMUVFew} & \checkvalue{62.72}{\GCLFTHIVFew} & \checkvalue{61.70}{\GCLFTBACEFew} & 53.76 \\
    & GPF-plus & \checkvalue{46.08}{\GCLFTPPIFew} & \checkvalue{53.35}{\GCLFTBBBPFew} & \checkvalue{50.36}{\GCLFTToxFew} & \checkvalue{52.66}{\GCLFTToxCastFew} & \checkvalue{50.08}{\GCLFTSIDERFew} & \checkvalue{60.79}{\GCLFTClinToxFew} & \checkvalue{49.53}{\GCLFTMUVFew} & \checkvalue{63.43}{\GCLFTHIVFew} & \checkvalue{61.72}{\GCLFTBACEFew} & 54.22\\
    \cline{2-12} 
    & SUPT$_{\text{hard}}$ & \checkvalue{50.48}{\GCLFTPPIFew} & \checkvalue{54.01}{\GCLFTBBBPFew} & \checkvalue{51.14}{\GCLFTToxFew} & \checkvalue{54.17}{\GCLFTToxCastFew} & \checkvalue{51.95}{\GCLFTSIDERFew} & \checkvalue{61.51}{\GCLFTClinToxFew} & \checkvalue{52.86}{\GCLFTMUVFew} & \checkvalue{63.82}{\GCLFTHIVFew} & \checkvalue{63.25}{\GCLFTBACEFew} & 55.91 \\
    & SUPT$_{\text{soft}}$ & \checkvalue{50.32}{\GCLFTPPIFew} & \checkvalue{53.96}{\GCLFTBBBPFew} & \checkvalue{52.21}{\GCLFTToxFew} & \checkvalue{54.20}{\GCLFTToxCastFew} & \checkvalue{52.00}{\GCLFTSIDERFew} & \checkvalue{61.62}{\GCLFTClinToxFew} & \checkvalue{52.94}{\GCLFTMUVFew} & \checkvalue{64.05}{\GCLFTHIVFew} & \checkvalue{63.61}{\GCLFTBACEFew} & 56.10 \\
    \midrule
    
    \multirow{5}{*}{\rotatebox{90}{EdgePred}} & FT & \EdgePredFTPPIFew & \EdgePredFTBBBPFew & \EdgePredFTToxFew & \EdgePredFTToxCastFew & \EdgePredFTSIDERFew & \EdgePredFTClinToxFew & \EdgePredFTMUVFew & \EdgePredFTHIVFew & \EdgePredFTBACEFew & 57.18  \\
    & GPF & \checkvalue{52.79}{\EdgePredFTPPIFew} & \checkvalue{51.38}{\EdgePredFTBBBPFew} & \checkvalue{64.00}{\EdgePredFTToxFew} & \checkvalue{57.10}{\EdgePredFTToxCastFew} & \checkvalue{49.90}{\EdgePredFTSIDERFew} & \checkvalue{67.82}{\EdgePredFTClinToxFew} & \checkvalue{67.58}{\EdgePredFTMUVFew} & \checkvalue{62.32}{\EdgePredFTHIVFew} & \checkvalue{62.03}{\EdgePredFTBACEFew} & 59.44 \\
    & GPF-plus & \checkvalue{52.88}{\EdgePredFTPPIFew} & \checkvalue{54.05}{\EdgePredFTBBBPFew} & \checkvalue{64.41}{\EdgePredFTToxFew} & \checkvalue{57.33}{\EdgePredFTToxCastFew} & \checkvalue{50.18}{\EdgePredFTSIDERFew} & \checkvalue{64.51}{\EdgePredFTClinToxFew} & \checkvalue{67.51}{\EdgePredFTMUVFew} & \checkvalue{62.85}{\EdgePredFTHIVFew} & \checkvalue{60.84}{\EdgePredFTBACEFew} & 59.40 \\
    \cline{2-12} 
    & SUPT$_{\text{hard}}$ & \checkvalue{57.42}{\EdgePredFTPPIFew} & \checkvalue{54.14}{\EdgePredFTBBBPFew} & \checkvalue{64.90}{\EdgePredFTToxFew} & \checkvalue{57.90}{\EdgePredFTToxCastFew} & \checkvalue{52.56}{\EdgePredFTSIDERFew} & \checkvalue{67.95}{\EdgePredFTClinToxFew} & \checkvalue{68.39}{\EdgePredFTMUVFew} & \checkvalue{63.44}{\EdgePredFTHIVFew} & \checkvalue{62.12}{\EdgePredFTBACEFew} & 60.98 \\
    & SUPT$_{\text{soft}}$ & \checkvalue{57.26}{\EdgePredFTPPIFew} & \checkvalue{54.36}{\EdgePredFTBBBPFew} & \checkvalue{65.19}{\EdgePredFTToxFew} & \checkvalue{57.88}{\EdgePredFTToxCastFew} & \checkvalue{52.45}{\EdgePredFTSIDERFew} & \checkvalue{67.70}{\EdgePredFTClinToxFew} & \checkvalue{68.73}{\EdgePredFTMUVFew} & \checkvalue{63.25}{\EdgePredFTHIVFew} & \checkvalue{61.70}{\EdgePredFTBACEFew} & 60.95 \\
    \bottomrule
    \end{tabular}
    \vskip 0.1in
    \caption{Performance measured by test ROC-AUC in 50-shot scenarios, with superior performances to FT highlighted in blue and others highlighted in red.
    The average values obtained from experiments repeated with 20 random seeds are reported. }
    \label{tab:few}
    \vskip -0.1in
\end{table*}

Given a pre-trained GNN $f_\theta$ and a task-specific projection head $h_\phi$, Fine-Tuning (FT) involves adjusting the parameters of both $\theta$ and $\phi$. 
In contrast, prompt tuning methods such as GPF, GPF-plus, and SUPT, entail tuning the parameters of $\psi$ for the prompt function $g_\psi$ as well as $\phi$, while maintaining $\theta$ as fixed.
In our experiments, we selected the projection head from an identical range of MLP layer counts as those used in GPF and GPF-plus.
The hyperparameter $k$ of SUPT in Equation (\ref{eq:SUPT}), representing the number of subgraphs, was chosen from the set \{1, 2, 3, 4, 5\}. 
The pooling ratio $r$ in Equation (\ref{eq:SUPT_hard}) for SUPT$_\text{hard}$ was selected from \{0.2, 0.4, 0.6\}.
We simply set the number of hops $m$ to 1; thereby $\alpha$ in Equation (\ref{eq:SUPT}) is calculated using GCN.
The Adam optimizer with weight decay \cite{DBLP:journals/corr/KingmaB14}, as utilized by \citeauthor{fang2023universal} and \citeauthor{hu2020pretraining}, was also employed for tuning SUPT. 
Additional information regarding the settings of the hyperparameters is provided in the appendix.
To ensure robustness and reliability of experimental results, we report the average performance from 20 random seeds. 
Consequently, the performance for GPF and GPF-plus reported in this study may deviate from those previously published, which were contingent upon evaluations conducted with 5 random seeds.
In this study, all experiments were conducted using an RTX 8000.
The full table including comparisons with other pre-training specific prompting methods such as GPPT \cite{sun2022gppt} and GraphPrompt \cite{liu2023graphprompt}, is detailed in the appendix. 
This is because our primary focus lies on the comparison of universal prompt tuning methods.

\subsection{Experiment Results}
In the comparison presented in Table \ref{tab:overall}, it is observed that all prompt tuning methods generally exhibit performance comparable to that of FT, despite having a significantly fewer number of tunable parameters.

\paragraph{\textbf{Full-shot scenarios}}
The experimental results in full-shot scenarios presented in Table \ref{tab:full} reveal that the SUPT models, based on AttrMasking, ContextPred, and EdgePred, consistently outperform FT across all cases. 
Conversely, GPF and GPF-plus models do not always exhibit superior performance. 
Specifically, SUPT$_\text{soft}$, leveraging Infomax, surpasses FT in every dataset, while SUPT$_\text{hard}$ exceeds FT in all but one dataset. 
In experiments employing GCL, both SUPT$_\text{soft}$ and SUPT$_\text{hard}$ underperform relative to FT in two datasets, yet they demonstrate superior performance in the majority of the remaining datasets.
Out of 45 experiments conducted across 9 datasets and 5 pre-training strategies, GPF and GPF-plus outperform FT in 27 cases, whereas SUPT$_\text{hard}$ and SUPT$_\text{soft}$ do so in 42 and 43 cases, respectively. 
On average, GPF and GPF-plus exhibit performance improvements of 0.91 and 1.06 over FT, respectively. In contrast, SUPT$_\text{hard}$ and SUPT$_\text{soft}$ show more notable enhancements, with average performance gains of 1.85 and 2.05, respectively.

\begin{table*}[t]
    \centering
    \vskip 0.15in
    \resizebox{\textwidth}{!}{
    \begin{tabular}{lc|ccccccccc}
    \toprule
        & $k$ & PPI &BBBP& Tox21 & ToxCast & SIDER & ClinTox & MUV & HIV & BACE \\
        \midrule
        \multirow{3}{*}{SUPT$_\text{hard}$} & $1$ & $7.94 \times 10^{-5}$ & $8.60 \times 10^{-5}$ & $7.67 \times 10^{-5}$ &$9.13 \times 10^{-5}$  &$7.65 \times 10^{-5}$ &$8.80 \times 10^{-5}$ &$7.73 \times 10^{-5}$ &$7.80 \times 10^{-5}$ &$7.54 \times 10^{-5}$ \\
        & $3$ & $1.94 \times 10^{-4}$ & $1.86 \times 10^{-4}$ & $2.01 \times 10^{-4}$ &$1.85\times 10^{-4}$  &$1.86 \times 10^{-4}$ &$1.87 \times 10^{-4}$ &$2.41 \times 10^{-4}$ &$1.86 \times 10^{-4}$ &$2.12 \times 10^{-4}$ \\
        & $5$ & $3.01 \times 10^{-4}$ & $3.24 \times 10^{-4}$ & $2.94 \times 10^{-4}$ &$3.22 \times 10^{-4}$  &$3.33 \times 10^{-4}$ &$2.94 \times 10^{-4}$ &$2.93 \times 10^{-4}$ &$3.42 \times 10^{-4}$ &$2.92 \times 10^{-4}$ \\
        \midrule
        \multirow{3}{*}{SUPT$_\text{soft}$}& $1$ & $2.11 \times 10^{-5}$ & $2.26 \times 10^{-5}$ & $2.24 \times 10^{-5}$ &$2.28 \times 10^{-5}$  &$1.92 \times 10^{-5}$ &$2.11 \times 10^{-5}$ &$2.19 \times 10^{-5}$ & $2.19 \times 10^{-5}$ &$1.95 \times 10^{-5}$ \\
        & $3$ & $2.35 \times 10^{-5}$ & $2.27 \times 10^{-5}$ & $2.30 \times 10^{-5}$ &$2.38 \times 10^{-5}$  &$2.30 \times 10^{-5}$ &$2.47 \times 10^{-5}$ &$2.21 \times 10^{-5}$ &$2.26 \times 10^{-5}$ &$1.94 \times 10^{-5}$ \\
        & $5$ & $2.42 \times 10^{-5}$ & $2.28 \times 10^{-5}$ & $2.46 \times 10^{-5}$ &$2.45 \times 10^{-5}$  &$2.30 \times 10^{-5}$ &$2.09 \times 10^{-5}$ &$2.24 \times 10^{-5}$ &$2.26 \times 10^{-5}$ &$1.94 \times 10^{-5}$ \\
    \bottomrule
    \end{tabular}
    }
    \vskip 0.1in
    \caption{Inference time (seconds) per graph for adding prompts $g_\psi(\mathcal{G})$, averaged over 1,000 repetitions, shows that SUPT$_\text{hard}$ experiences a more pronounced increase in processing time with the addition of subgraphs compared to SUPT$_\text{soft}$.}
    \label{tab:time}
    \vskip -0.1in
\end{table*}

\paragraph{\textbf{Few-shot scenarios}}
As depicted in Table \ref{tab:few}, the results from 50-shot experiments across 45 different settings are reported. 
Both SUPT$_\text{hard}$ and SUPT$_\text{soft}$ demonstrate superior performance over FT across all datasets in experiments based on Infomax and AttrMasking. 
In the case of ContextPred and EdgePred, they outperform FT in all but one dataset for each strategy, and for GCL, they exceed FT in all but two datasets.
Generally, SUPT demonstrates a consistent ability to outperform FT across a variety of pre-training strategies. 
On the other hand, GPF falls short of surpassing FT in four datasets within the GCL strategy. 
Out of the total 45 experiments, GPF outperforms FT in 33 cases, GPF-plus in 35 cases, and both SUPT$_\text{hard}$ and SUPT$_\text{soft}$ in 41 cases. 
On average, GPF and GPF-plus show performance improvements of 1.89 and 2.00 over FT, respectively, while SUPT$_\text{hard}$ and SUPT$_\text{soft}$ exhibit even higher average performance gains of 3.73 and 3.67, respectively. 
This indicates that all four prompt tuning methods yield greater performance enhancements in the 50-shot setting compared to the full-shot setting.

\section{Discussion}
\label{sec:discussion}

\paragraph{\textbf{Efficiency in Parameters and Computation}}
As reported in Table \ref{tab:overall}, the number of learning parameters for prompt tuning methods (denoted as $\psi$ in Equation (\ref{eq:obj_prompt})) is significantly lower than those required for fine-tuning methods (represented by $\theta$ in Equation (\ref{eq:obj_fine})), with all being less than 1\% of the latter. 
Notably, SUPT necessitates more parameters than GPF but fewer than GPF-plus. 
While expanding the number of subgraphs ($k$ in Equation (\ref{eq:SUPT})) in SUPT might imply a higher demand for parameters, empirical evidence demonstrates that SUPT is capable of surpassing the performance of GPF-plus without necessitating an increase in the number of subgraphs.
This underscores the advantage of employing more intricately patterned prompts over uniformly applied or node-type-specific ones in capturing the complex and diverse nature of graph data.

The limitation of SUPT lies in the additional computational step required to calculate $\alpha$ in Equation (\ref{eq:SUPT}), which involves operations such as $\tilde{D}^{-\frac{1}{2}}\tilde{A}\tilde{D}^{-\frac{1}{2}}XW$.
This step has a computational complexity of $\mathcal{O}(|\mathcal{E}|d + |\mathcal{V}|dk)$, potentially impacting practical inference times.
Testing with the BBBP dataset, where each graph's prompt addition was repeated 1,000 times, showed an average added time of $2.27 \times 10^{-5}$ seconds per graph for SUPT$_\text{soft}$ and $1.86 \times 10^{-4}$ seconds for SUPT$_\text{hard}$, with the number of subgraphs $k=3$ as reported in Table \ref{tab:time}. 
The computational time for SUPT$_\text{hard}$ is longer due to the top-rank operation in Equation (\ref{eq:SUPT_hard}).
Also, SUPT$_\text{hard}$ exhibits a greater escalation in processing time as the number of subgraphs increases, in contrast to SUPT$_\text{soft}$.
Despite this, the increase in inference time is negligible, suggesting that the practical implications are manageable and the benefits of SUPT's enhanced performance are significant.

\paragraph{\textbf{Patterns of Subgraph Selection}}
In Figure \ref{fig:SUPT}, the subgraphs of SUPT are depicted as if they are composed solely of adjacent nodes for the sake of a conceptual comparison, yet there are no technical constraints mandating such a configuration. 
However, due to the use of GCN or SGC in calculating scores for each prompt basis as per Equation (\ref{eq:SUPT}), adjacent nodes tend to exhibit similar scoring patterns for each basis. 
This phenomenon aligns with numerous studies asserting that GNNs act like low-pass filters or possess smoothing effects \cite{xu2018representation,nt2019revisiting,oono2020graph}. 
Consequently, even without employing specific techniques, nodes in proximity tend to be grouped into the same subgraph, somewhat akin to the illustration in Figure \ref{fig:SUPT}. 
Furthermore, inspired by the DiffPool \cite{ying2018hierarchical}, which also influenced the development of SUPT$_\text{soft}$, an auxiliary loss $\mathcal{L}_{LP}$ in the form of link prediction can be incorporated to enhance the model's performance:
\begin{equation}
    \mathcal{L}_{LP} = \| A, \alpha \alpha^\top \|_F
\end{equation}
where $\|\cdot\|_F$ denotes the Frobenius norm.

\section{Conclusion}
\label{sec:conclusion}
In this study, we introduced the Subgraph-level Universal Prompt Tuning (SUPT) approach, focusing on enhancing the prompt tuning process for graph neural networks by incorporating prompt features at the subgraph level. 
This method maintains the flexibility of prompt tuning across various pre-training strategies while requiring significantly fewer tuning parameters than traditional fine-tuning methods. 
Our experimental results demonstrate the effectiveness of SUPT, showing notable improvements in performance in both full-shot and few-shot learning scenarios, outperforming existing methods in the majority of our tests. 
These findings suggest that SUPT is a viable and efficient alternative for adapting pre-trained graph neural networks for diverse downstream tasks, although the exploration of simple prompts' ability to capture complex graph contexts warrants further investigation. 
The study contributes to the broader machine learning field by offering insights into the potential of prompt tuning in graph-based applications, highlighting the balance between efficiency and effectiveness in model adaptation strategies.


\newpage

\bibliographystyle{ACM-Reference-Format}
\bibliography{references}

\newpage
\appendix
\section{Proof for Theorem 1}

\theoremstyle{plain}
\newtheorem{theorem1}{Theorem}
\theoremstyle{plain}
\newtheorem{proposition1}{Proposition}

\begin{theorem1}[\textbf{Universal Capability of SUPT}]
Let $ f $ be a pre-trained GNN model and $ \mathcal{G} : (X,A) $ be an input graph with sets of nodes and edges, $\mathcal{V}$ and $\mathcal{E}$. 
Given any prompting function $ \psi_t(\cdot) $, if a prompted graph $ \hat{\mathcal{G}} : (\hat{X} \in \mathbb{X}, \hat{A} \in \mathbb{A}) $ is in the candidate space $ \mathcal{G}^* = \Psi_t(\mathcal{G}) $, then there exists a vector $ \hat{p} $ such that for any non-empty subset of nodes $ \mathcal{S} $ (i.e., $ \mathcal{S} \subseteq \mathcal{V} $ and $ |\mathcal{S}| \geq 1 $), the updated features can be defined as:
\[ 
X' = 
\begin{cases} 
X_s + \hat{p} & \text{if } s \in \mathcal{S} \\
X_s & \text{if } s \notin \mathcal{S} 
\end{cases}
\]
such that there exists $X' $ satisfying:
\[
f(X',A) = f(\hat{X},\hat{A})
\]
\label{th1:universality}
\end{theorem1}
To demonstrate Theorem \ref{th1:universality}, we detail the architecture of the pre-trained GNN $f_\theta$. 
For ease of analysis, we initially consider $f_\theta$ to be a single-layer Graph Isomorphism Network (GIN) \cite{xu2018how}, incorporating a linear transformation, as done in \citeauthor{fang2023universal}.
The embedding is calculated through the pre-trained GNN model $f_\theta$ as:
\begin{align}    
    Z &= (A+ (1+\epsilon)\cdot I) \cdot X \cdot W \\
    z_\mathcal{G} &= \sum_{v_i \in \mathcal{V}} z_i
\end{align}
where a parameter $\epsilon$ and $W$ for linear projection are elements of parameter set $\theta$.
They have been pre-trained and then fixed in prompt tuning process.
With these assumptions in place, we demonstrate that adding a prompt feature vector to a subset of nodes can theoretically be equivalent to the Graph Prompt Feature (GPF) approach \cite{fang2023universal}, which incorporates prompts into all nodes.
Theorem \ref{th1:universality} is equivalent to Proposition \ref{th1:proposition}.

\begin{proposition1}
Let $ f $ be a pre-trained GNN model and $ \mathcal{G} : (X,A) $ be an input graph with sets of nodes and edges, $\mathcal{V}$ and $\mathcal{E}$. 
There exists a vector $ \hat{p} $ such that for any non-empty subset of nodes $ \mathcal{S} $ (i.e., $ \mathcal{S} \subseteq \mathcal{V} $ and $ |\mathcal{S}| \geq 1 $), the updated features can be defined as:
\[ 
X' = 
\begin{cases} 
X_s + \hat{p} & \text{if } s \in \mathcal{S} \\
X_s & \text{if } s \notin \mathcal{S} 
\end{cases}
\]
such that there exists $X' $ satisfying:
\[
f(X',A) = f(X+p, A)
\]
where $p$ is a uniform feature vector added to all nodes in $X$.
\label{th1:proposition}
\end{proposition1}

\begin{proof}
    For $p = [\rho_1,...,\rho_F]\in \mathbb{R}^{1 \times F}$, we have:
    \begin{align}
        Z_p &= (A+ (1+\epsilon)\cdot I) \cdot (X + [1]^N\cdot p) \cdot W \\
        &= (A+ (1+\epsilon)\cdot I) \cdot X \cdot W + (A+ (1+\epsilon)\cdot I) \cdot [1]^N\cdot p \cdot W \\
        &= Z + (A+ (1+\epsilon)\cdot I) \cdot [1]^N\cdot p \cdot W\\
        &= Z+ [d_i+1+\epsilon]^N \cdot p \cdot W
    \end{align}
    where $[1]^N \in \mathbb{R}^{N \times 1}$ and $[d_i+1+\epsilon]^N \in \mathbb{R}^{N \times 1}$ are column vectors and $d_i$ denotes the degree of node $v_i$.

    For a subset of nodes $\mathcal{S}$ and SUPT $\hat{p} = [\hat{\rho_1},...,\hat{\rho_F}]\in \mathbb{R}^{1 \times F}$, we have:
    \begin{align}
        Z_{\hat{p}} &= (A+ (1+\epsilon)\cdot I) \cdot (X + [\mathbf{1}_\mathcal{S}(v)]^N\cdot \hat{p}) \cdot W \\
        &= (A+ (1+\epsilon)\cdot I) \cdot X \cdot W + (A+ (1+\epsilon)\cdot I) \cdot [\mathbf{1}_\mathcal{S}(v)]^N\cdot \hat{p} \cdot W \\
        &= Z + (A+ (1+\epsilon)\cdot I) \cdot [\mathbf{1}_\mathcal{S}(v)]^N\cdot \hat{p} \cdot W
    \end{align}
    where $\mathbf{1}_\mathcal{S}(v)$ denotes the indicator function defined to be 1 if $v\in \mathcal{S}$ and 0 otherwise.

    To obtain the identical representation $z_\mathcal{G}$, we have:
    \begin{equation}
        z_{\mathcal{G},\hat{p}} = z_{\mathcal{G},p} \rightarrow \text{Sum}(Z_{\hat{p}}) =\text{Sum}(Z_{p})
    \end{equation}
    where $\text{Sum}(M) = \sum_i m_i$ denotes the summation of each row of the matrix.
    We simplify the above equation as:
    \begin{align}
        &z_{\mathcal{G},\hat{p}} = z_{\mathcal{G},p} \\
        \rightarrow& \text{Sum}(Z_{\hat{p}}) =\text{Sum}(Z_{p})\\
        \rightarrow&  \text{Sum}( Z + (A+ (1+\epsilon)\cdot I) \cdot [\mathbf{1}_\mathcal{S}(v)]^N\cdot \hat{p} \cdot W) \\
        &=\text{Sum}(Z+ [d_i+1+\epsilon]^N \cdot p \cdot W)\\
        \rightarrow&  \text{Sum}((A+ (1+\epsilon)\cdot I) \cdot [\mathbf{1}_\mathcal{S}(v)]^N\cdot \hat{p} \cdot W) \\
        &=\text{Sum}([d_i+1+\epsilon]^N \cdot p \cdot W)
    \end{align}
    Then, we calculate $\Delta z_{\mathcal{G},p} = \text{Sum}([d_i+1+\epsilon]^N \cdot p \cdot W) \in \mathbb{R}^{d'}$ as:
    \begin{align}
        \Delta z_{\mathcal{G},p}^i &= \sum_{j=1}^{d}\sum_{k=1}^N (d_i+1+\epsilon) \cdot \rho_j \cdot W_{j,i} \\
        &=\sum_{j=1}^{d} (D+N+N\cdot\epsilon)\cdot \rho_j \cdot W_{j,i}
    \end{align}
    where $\rho_j$ for $j\in [1,d]$ denotes $j$-th parameter of GPF feature vector, $\Delta z_{\mathcal{G},p}^i$ denotes $i$-dimension in $\Delta z_{\mathcal{G},p}^i$, and $D=\sum_{k=1}^N d_k$ is the sum of degree of all nodes.
    Let $(A+ (1+\epsilon)\cdot I) \cdot [\mathbf{1}_\mathcal{S}(v)]^N\cdot \hat{p}=B\in\mathbb{R}^{N \times d}$, then we have:
    \begin{equation}
        \Delta z_{\mathcal{G},\hat{p}}^i = \sum_{j=1}^d (\sum_{k=1}^N \beta_{k,j})\cdot W_{j,i}
    \end{equation}
    where $\beta_{k,j}$ for $k\in[1,N], j\in[1,d]$ denotes the learnable parameter in $B$, which is the parameter of SUPT prompt.
    Then we have:
    \begin{align}
        &z_{\mathcal{G},\hat{p}}^i = z_{\mathcal{G},p}^i , \text{ for every } i \in [1, d']\\
        \rightarrow& \Delta z_{\mathcal{G},\hat{p}}^i = \Delta z_{\mathcal{G},p}^i \\
        \rightarrow& \sum_{k=1}^N \beta_{k,j} = (D+N+N\cdot \epsilon) \alpha_j, \ \  j \in [1,d]
    \end{align}
    Therefore, there exists a SUPT prompt $\hat{p}$ that satisfies the above conditions to be equivalent to a GPF prompt $p$ for the pre-trained GNN model $f_\theta$. 
\end{proof}

Theorem \ref{th1:universality} and Proposition \ref{th1:proposition} imply that SUPT can, in theory, achieve equivalence with GPF, and thus be equivalent to any prompting function.
\citeauthor{fang2023universal} proved that GPF satisfies Proposition \ref{th1:proposition_GPF1} and \ref{th1:proposition_GPF2} below.

\begin{proposition1}
    Given a pre-trained GNN model $f$, an input graph $\mathcal{G}:(X,A)$, for any graph-level transformation $g:\mathbb{G}\rightarrow\mathbb{G}$, there exists a GPF extra feature vector $p$ that satisfies:
    \begin{equation}
        f(X+p,A) = f(g(X,A))
    \end{equation}
    \label{th1:proposition_GPF1}
\end{proposition1}

\begin{proposition1}
    Given an input graph $\mathcal{G}:(X,A)$, an arbitrary graph-level transformation $g:\mathbb{G}\rightarrow\mathbb{G}$ can be decoupled to a series of following transformations:
    \begin{itemize}
        \item \textbf{Feature transformations.} Modifying the node features and generating the new feature matrix $X'=g_{ft}(X)$.
        \item \textbf{Link transformations.} Adding or removing edges and generating the new adjacency matrix $A'=g_{lt}(A)$
        \item \textbf{Isolated component transformations.} Adding or removing isolated components (sub-graphs) and generating the new adjacency matrix and feature matrix $X',A' = g_{ict}(X,A)$.
    \end{itemize}
    \label{th1:proposition_GPF2}
\end{proposition1}

\section{Details on Training and Hyperparameters}
For SUPT, as defined in Equation (\ref{eq:SUPT}), the hyperparameter $k$, which denotes the number of subgraphs, was selected from the set \{1, 2, 3, 4, 5\}. 
In the case of SUPT$_\text{hard}$, the pooling ratio $r$, as outlined in Equation (\ref{eq:SUPT_hard}), was chosen from the options \{0.2, 0.4, 0.6\}. 
We fixed the number of hops $m$ at 1, leading to the calculation of $\alpha$ in Equation (\ref{eq:SUPT}) via a GCN. 
The weight decay parameters were determined from the set \{1e-3, 1e-4, 1e-5\}, learning rates from \{1e-3, 5e-4, 1e-4\}, the number of layers of projection head from \{1, 2, 3, 4\}, the global pooling from \{\textit{sum}, \textit{mean}\}, and the number of epochs from \{50, 100, 300\}.

\begin{table*}[t]
    \centering
    \vskip 0.15in
    \resizebox{0.75\textwidth}{!}{
    \begin{tabular}{ll|ccccccccc|r}
        \toprule
         & Tuning & PPI & BBBP & Tox21 & ToxCast & SIDER & ClinTox & MUV & HIV & BACE & Avg. \\
        \midrule 
        
        \multirow{5}{*}{\rotatebox{90}{Infomax}} & FT & \InfomaxFTPPIFull & \InfomaxFTBBBPFull & \InfomaxFTToxFull & \InfomaxFTToxCastFull & \InfomaxFTSIDERFull & \InfomaxFTClinToxFull & \InfomaxFTMUVFull & \InfomaxFTHIVFull & \InfomaxFTBACEFull & 72.94 \\
        & GPF & \checkvalue{78.87}{\InfomaxFTPPIFull} & \checkvalue{66.58}{\InfomaxFTBBBPFull} & \checkvalue{78.13}{\InfomaxFTToxFull} & \checkvalue{65.96}{\InfomaxFTToxCastFull} & \checkvalue{65.64}{\InfomaxFTSIDERFull} & \checkvalue{74.14}{\InfomaxFTClinToxFull} & \checkvalue{80.39}{\InfomaxFTMUVFull} & \checkvalue{76.67}{\InfomaxFTHIVFull} & \checkvalue{83.24}{\InfomaxFTBACEFull} & 74.40 \\
        & GPF-plus & \checkvalue{78.85}{\InfomaxFTPPIFull} & \checkvalue{67.03}{\InfomaxFTBBBPFull} & \checkvalue{78.50}{\InfomaxFTToxFull} & \checkvalue{65.83}{\InfomaxFTToxCastFull} & \checkvalue{65.89}{\InfomaxFTSIDERFull} & \checkvalue{73.61}{\InfomaxFTClinToxFull} & \checkvalue{79.94}{\InfomaxFTMUVFull} & \checkvalue{75.60}{\InfomaxFTHIVFull} & \checkvalue{82.96}{\InfomaxFTBACEFull} & 74.25 \\
        \cline{2-12} 
        & SUPT$_{\text{hard}}$ & \checkvalue{79.64}{\InfomaxFTPPIFull} & \checkvalue{67.29}{\InfomaxFTBBBPFull} & \checkvalue{79.14}{\InfomaxFTToxFull} & \checkvalue{66.48}{\InfomaxFTToxCastFull} & \checkvalue{65.78}{\InfomaxFTSIDERFull} & \checkvalue{75.65}{\InfomaxFTClinToxFull} & \checkvalue{81.51}{\InfomaxFTMUVFull} & \checkvalue{77.96}{\InfomaxFTHIVFull} & \checkvalue{84.05}{\InfomaxFTBACEFull} & 75.28 \\
        & SUPT$_{\text{soft}}$ & \checkvalue{79.45}{\InfomaxFTPPIFull} & \checkvalue{67.63}{\InfomaxFTBBBPFull} & \checkvalue{78.98}{\InfomaxFTToxFull} & \checkvalue{66.53}{\InfomaxFTToxCastFull} & \checkvalue{65.88}{\InfomaxFTSIDERFull} & \checkvalue{76.08}{\InfomaxFTClinToxFull} & \checkvalue{81.59}{\InfomaxFTMUVFull} & \checkvalue{77.94}{\InfomaxFTHIVFull} & \checkvalue{84.01}{\InfomaxFTBACEFull} & 75.34 \\
        \midrule
        
        \multirow{5}{*}{\rotatebox{90}{AttrMasking}} & FT & \AttrMaskingFTPPIFull & \AttrMaskingFTBBBPFull & \AttrMaskingFTToxFull & \AttrMaskingFTToxCastFull & \AttrMaskingFTSIDERFull & \AttrMaskingFTClinToxFull & \AttrMaskingFTMUVFull & \AttrMaskingFTHIVFull & \AttrMaskingFTBACEFull & 73.97 \\
        & GPF & \checkvalue{81.47}{\AttrMaskingFTPPIFull} & \checkvalue{67.49}{\AttrMaskingFTBBBPFull} & \checkvalue{78.66}{\AttrMaskingFTToxFull} & \checkvalue{66.09}{\AttrMaskingFTToxCastFull} & \checkvalue{68.19}{\AttrMaskingFTSIDERFull} & \checkvalue{72.72}{\AttrMaskingFTClinToxFull} & \checkvalue{81.44}{\AttrMaskingFTMUVFull} & \checkvalue{77.09}{\AttrMaskingFTHIVFull} & \checkvalue{84.77}{\AttrMaskingFTBACEFull} & 75.32 \\
        & GPF-plus & \checkvalue{81.43}{\AttrMaskingFTPPIFull} & \checkvalue{67.32}{\AttrMaskingFTBBBPFull} & \checkvalue{78.83}{\AttrMaskingFTToxFull} & \checkvalue{66.53}{\AttrMaskingFTToxCastFull} & \checkvalue{68.53}{\AttrMaskingFTSIDERFull} & \checkvalue{74.17}{\AttrMaskingFTClinToxFull} & \checkvalue{81.27}{\AttrMaskingFTMUVFull} & \checkvalue{77.64}{\AttrMaskingFTHIVFull} & \checkvalue{85.01}{\AttrMaskingFTBACEFull} & 75.64 \\
        \cline{2-12} 
        & SUPT$_{\text{hard}}$ & \checkvalue{81.90}{\AttrMaskingFTPPIFull} & \checkvalue{68.39}{\AttrMaskingFTBBBPFull} & \checkvalue{79.50}{\AttrMaskingFTToxFull} & \checkvalue{66.79}{\AttrMaskingFTToxCastFull} & \checkvalue{68.82}{\AttrMaskingFTSIDERFull} & \checkvalue{75.25}{\AttrMaskingFTClinToxFull} & \checkvalue{81.81}{\AttrMaskingFTMUVFull} & \checkvalue{79.10}{\AttrMaskingFTHIVFull} & \checkvalue{84.98}{\AttrMaskingFTBACEFull} & 76.28\\
        & SUPT$_{\text{soft}}$& \checkvalue{81.91}{\AttrMaskingFTPPIFull} & \checkvalue{68.37}{\AttrMaskingFTBBBPFull} & \checkvalue{79.73}{\AttrMaskingFTToxFull} & \checkvalue{67.02}{\AttrMaskingFTToxCastFull} & \checkvalue{68.68}{\AttrMaskingFTSIDERFull} & \checkvalue{77.36}{\AttrMaskingFTClinToxFull} & \checkvalue{81.93}{\AttrMaskingFTMUVFull} & \checkvalue{79.15}{\AttrMaskingFTHIVFull} & \checkvalue{85.22}{\AttrMaskingFTBACEFull} & 76.60 \\
        \midrule
        
        \multirow{5}{*}{\rotatebox{90}{ContextPred}} & FT & \ContextPredFTPPIFull & \ContextPredFTBBBPFull & \ContextPredFTToxFull & \ContextPredFTToxCastFull & \ContextPredFTSIDERFull & \ContextPredFTClinToxFull & \ContextPredFTMUVFull & \ContextPredFTHIVFull & \ContextPredFTBACEFull & 74.53 \\
        & GPF & \checkvalue{80.21}{\ContextPredFTPPIFull} & \checkvalue{69.04}{\ContextPredFTBBBPFull} & \checkvalue{79.90}{\ContextPredFTToxFull} & \checkvalue{67.68}{\ContextPredFTToxCastFull} & \checkvalue{65.31}{\ContextPredFTSIDERFull} & \checkvalue{73.15}{\ContextPredFTClinToxFull} & \checkvalue{84.22}{\ContextPredFTMUVFull} & \checkvalue{77.77}{\ContextPredFTHIVFull} & \checkvalue{85.03}{\ContextPredFTBACEFull} & 75.81 \\
        & GPF-plus & \checkvalue{80.28}{\ContextPredFTPPIFull} & \checkvalue{68.59}{\ContextPredFTBBBPFull} & \checkvalue{79.95}{\ContextPredFTToxFull} & \checkvalue{67.72}{\ContextPredFTToxCastFull} & \checkvalue{66.32}{\ContextPredFTSIDERFull} & \checkvalue{72.73}{\ContextPredFTClinToxFull} & \checkvalue{84.40}{\ContextPredFTMUVFull} & \checkvalue{78.10}{\ContextPredFTHIVFull} & \checkvalue{84.50}{\ContextPredFTBACEFull} & 75.84 \\
        \cline{2-12} 
        & SUPT$_{\text{hard}}$ & \checkvalue{80.85}{\ContextPredFTPPIFull} & \checkvalue{70.18}{\ContextPredFTBBBPFull} & \checkvalue{80.00}{\ContextPredFTToxFull} & \checkvalue{68.40}{\ContextPredFTToxCastFull} & \checkvalue{66.37}{\ContextPredFTSIDERFull} & \checkvalue{74.90}{\ContextPredFTClinToxFull} & \checkvalue{84.54}{\ContextPredFTMUVFull} & \checkvalue{79.42}{\ContextPredFTHIVFull} & \checkvalue{85.21}{\ContextPredFTBACEFull} & 76.65\\
        & SUPT$_{\text{soft}}$ & \checkvalue{80.86}{\ContextPredFTPPIFull} & \checkvalue{70.06}{\ContextPredFTBBBPFull} & \checkvalue{80.12}{\ContextPredFTToxFull} & \checkvalue{68.50}{\ContextPredFTToxCastFull} & \checkvalue{66.48}{\ContextPredFTSIDERFull} & \checkvalue{75.55}{\ContextPredFTClinToxFull} & \checkvalue{84.67}{\ContextPredFTMUVFull} & \checkvalue{79.39}{\ContextPredFTHIVFull} & \checkvalue{85.27}{\ContextPredFTBACEFull} & 76.77\\

        \midrule
        \multirow{5}{*}{\rotatebox{90}{GCL}} & FT & \GCLFTPPIFull & \GCLFTBBBPFull & \GCLFTToxFull & \GCLFTToxCastFull & \GCLFTSIDERFull & \GCLFTClinToxFull & \GCLFTMUVFull & \GCLFTHIVFull & \GCLFTBACEFull & 70.27 \\
        & GPF & \checkvalue{67.91}{\GCLFTPPIFull} & \checkvalue{68.11}{\GCLFTBBBPFull} & \checkvalue{72.15}{\GCLFTToxFull} & \checkvalue{60.73}{\GCLFTToxCastFull} & \checkvalue{61.00}{\GCLFTSIDERFull} & \checkvalue{69.48}{\GCLFTClinToxFull} & \checkvalue{70.81}{\GCLFTMUVFull} & \checkvalue{75.12}{\GCLFTHIVFull} & \checkvalue{76.67}{\GCLFTBACEFull} & 69.11 \\
        & GPF-plus & \checkvalue{66.71}{\GCLFTPPIFull} & \checkvalue{67.77}{\GCLFTBBBPFull} & \checkvalue{73.06}{\GCLFTToxFull} & \checkvalue{61.47}{\GCLFTToxCastFull} & \checkvalue{61.57}{\GCLFTSIDERFull} & \checkvalue{68.89}{\GCLFTClinToxFull} & \checkvalue{70.54}{\GCLFTMUVFull} & \checkvalue{75.90}{\GCLFTHIVFull} & \checkvalue{77.84}{\GCLFTBACEFull} & 69.31 \\
        \cline{2-12} 
        & SUPT$_{\text{hard}}$ & \checkvalue{68.59}{\GCLFTPPIFull} & \checkvalue{69.68}{\GCLFTBBBPFull} & \checkvalue{73.44}{\GCLFTToxFull} & \checkvalue{63.35}{\GCLFTToxCastFull} & \checkvalue{62.51}{\GCLFTSIDERFull} & \checkvalue{69.47}{\GCLFTClinToxFull} & \checkvalue{70.64}{\GCLFTMUVFull} & \checkvalue{76.22}{\GCLFTHIVFull} & \checkvalue{79.17}{\GCLFTBACEFull} & 70.34 \\
        & SUPT$_{\text{soft}}$ & \checkvalue{69.57}{\GCLFTPPIFull} & \checkvalue{69.79}{\GCLFTBBBPFull} & \checkvalue{73.40}{\GCLFTToxFull} & \checkvalue{63.42}{\GCLFTToxCastFull} & \checkvalue{62.84}{\GCLFTSIDERFull} & \checkvalue{71.21}{\GCLFTClinToxFull} & \checkvalue{70.92}{\GCLFTMUVFull} & \checkvalue{76.44}{\GCLFTHIVFull} & \checkvalue{78.92}{\GCLFTBACEFull} & 70.72 \\
        \midrule
        
        \multirow{5}{*}{\rotatebox{90}{EdgePred}} & FT & \EdgePredFTPPIFull & \EdgePredFTBBBPFull & \EdgePredFTToxFull & \EdgePredFTToxCastFull & \EdgePredFTSIDERFull & \EdgePredFTClinToxFull & \EdgePredFTMUVFull & \EdgePredFTHIVFull & \EdgePredFTBACEFull & 72.72 \\
        & GPPT & \checkvalue{56.23}{\EdgePredFTPPIFull} & \checkvalue{64.13}{\EdgePredFTBBBPFull} & \checkvalue{66.41}{\EdgePredFTToxFull} & \checkvalue{60.34}{\EdgePredFTToxCastFull} & \checkvalue{54.86}{\EdgePredFTSIDERFull} & \checkvalue{59.81}{\EdgePredFTClinToxFull} & \checkvalue{63.05}{\EdgePredFTMUVFull} & \checkvalue{60.54}{\EdgePredFTHIVFull} & \checkvalue{70.85}{\EdgePredFTBACEFull} & 61.80 \\
        & GPPT$_{\text{w/o ol}}$ & \checkvalue{76.85}{\EdgePredFTPPIFull} & \checkvalue{69.43}{\EdgePredFTBBBPFull} & \checkvalue{78.91}{\EdgePredFTToxFull} & \checkvalue{64.86}{\EdgePredFTToxCastFull} & \checkvalue{60.94}{\EdgePredFTSIDERFull} & \checkvalue{62.15}{\EdgePredFTClinToxFull} & \checkvalue{82.06}{\EdgePredFTMUVFull} & \checkvalue{73.19}{\EdgePredFTHIVFull} & \checkvalue{70.31}{\EdgePredFTBACEFull} & 70.97 \\
        & GraphPrompt & \checkvalue{49.48}{\EdgePredFTPPIFull} & \checkvalue{69.29}{\EdgePredFTBBBPFull} & \checkvalue{68.09}{\EdgePredFTToxFull} & \checkvalue{60.54}{\EdgePredFTToxCastFull} & \checkvalue{58.71}{\EdgePredFTSIDERFull} & \checkvalue{55.37}{\EdgePredFTClinToxFull} & \checkvalue{62.35}{\EdgePredFTMUVFull} & \checkvalue{59.31}{\EdgePredFTHIVFull} & \checkvalue{67.70}{\EdgePredFTBACEFull} & 61.20 \\
        & GPF & \checkvalue{79.96}{\EdgePredFTPPIFull} & \checkvalue{69.36}{\EdgePredFTBBBPFull} & \checkvalue{79.69}{\EdgePredFTToxFull} & \checkvalue{65.70}{\EdgePredFTToxCastFull} & \checkvalue{66.71}{\EdgePredFTSIDERFull} & \checkvalue{67.92}{\EdgePredFTClinToxFull} & \checkvalue{82.61}{\EdgePredFTMUVFull} & \checkvalue{77.53}{\EdgePredFTHIVFull} & \checkvalue{79.76}{\EdgePredFTBACEFull} & 74.36 \\
        & GPF-plus & \checkvalue{79.88}{\EdgePredFTPPIFull} & \checkvalue{69.31}{\EdgePredFTBBBPFull} & \checkvalue{79.91}{\EdgePredFTToxFull} & \checkvalue{65.66}{\EdgePredFTToxCastFull} & \checkvalue{66.61}{\EdgePredFTSIDERFull} & \checkvalue{69.58}{\EdgePredFTClinToxFull} & \checkvalue{82.74}{\EdgePredFTMUVFull} & \checkvalue{77.77}{\EdgePredFTHIVFull} & \checkvalue{80.91}{\EdgePredFTBACEFull} & 74.71 \\
        \cline{2-12} 
        & SUPT$_{\text{hard}}$ & \checkvalue{80.20}{\EdgePredFTPPIFull} & \checkvalue{69.51}{\EdgePredFTBBBPFull} & \checkvalue{80.76}{\EdgePredFTToxFull} & \checkvalue{66.44}{\EdgePredFTToxCastFull} & \checkvalue{66.85}{\EdgePredFTSIDERFull} & \checkvalue{69.97}{\EdgePredFTClinToxFull} & \checkvalue{82.94}{\EdgePredFTMUVFull} & \checkvalue{77.79}{\EdgePredFTHIVFull} & \checkvalue{81.72}{\EdgePredFTBACEFull} & 75.13 \\
        & SUPT$_{\text{soft}}$ & \checkvalue{80.19}{\EdgePredFTPPIFull} & \checkvalue{69.93}{\EdgePredFTBBBPFull} & \checkvalue{80.90}{\EdgePredFTToxFull} & \checkvalue{66.65}{\EdgePredFTToxCastFull} & \checkvalue{66.95}{\EdgePredFTSIDERFull} & \checkvalue{70.41}{\EdgePredFTClinToxFull} & \checkvalue{82.95}{\EdgePredFTMUVFull} & \checkvalue{77.95}{\EdgePredFTHIVFull} & \checkvalue{81.44}{\EdgePredFTBACEFull} & 75.26 \\
        \bottomrule
    \end{tabular}
    }
    \vskip 0.1in
    \caption{Comparison of test ROC-AUC performance in full-shot scenarios across protein function prediction and molecular prediction benchmarks using various pre-training strategies, highlighted in blue for performances superior to fine-tuning (FT) and in red otherwise. }
    
    \vskip -0.1in
    \label{tab:appendix_full}
\end{table*}

\end{document}